\documentclass{article}
\usepackage{iclr2027_conference,times}
\usepackage[T1]{fontenc}
\usepackage{microtype,amsmath,amssymb,booktabs,tabularx,xcolor,graphicx}
\usepackage[hidelinks,pagebackref]{hyperref}

\usepackage{xurl,placeins}
\usepackage{comment}

\hypersetup{pdftitle={Gaussian Distillation of Flow Policies},pdfauthor={Youssef Saied, Ziyad Sheebaelhamd, Claire Vernade, François Fleuret}}

\title{Gaussian Distillation of Flow Policies}
\author{%
    \begin{tabular}{@{}p{0.48\textwidth}@{\hspace{0.04\textwidth}}p{0.48\textwidth}@{}}
        \begin{minipage}[t]{\linewidth}\vspace{0pt}\raggedright
            \textbf{Youssef Saied} \\
            \normalfont University of Geneva \\
            Geneva, Switzerland \\
            \texttt{youssef.saied@unige.ch}
        \end{minipage}
        &
        \begin{minipage}[t]{\linewidth}\vspace{0pt}\raggedright
            \textbf{Ziyad Sheebaelhamd} \\
            \normalfont University of Tübingen \\
            Tübingen, Germany
        \end{minipage}
        \\[2pt]
        \begin{minipage}[t]{\linewidth}\vspace{0pt}\raggedright
            \textbf{Claire Vernade} \\
            \normalfont University of Technology Nuremberg \\
            Nuremberg, Germany
        \end{minipage}
        &
        \begin{minipage}[t]{\linewidth}\vspace{0pt}\raggedright
            \textbf{François Fleuret} \\
            \normalfont University of Geneva and Meta FAIR \\
            Geneva, Switzerland
        \end{minipage}
    \end{tabular}
}
\iclrfinalcopy
\renewcommand{\headrulewidth}{0pt}
\begin{document}
\raggedbottom
\setlength{\textfloatsep}{9pt plus 2pt minus 2pt}
\setlength{\floatsep}{7pt plus 2pt minus 2pt}
\setlength{\intextsep}{6pt plus 2pt minus 2pt}
\begingroup
\setlength{\tabcolsep}{0pt}
\vspace*{-24pt}
\maketitle
\endgroup
\lhead{}
\vspace{-20pt}

\begin{abstract}
    Flow policies are among the strongest robot controllers and are well suited to modelling multimodal action distributions.
    Our analysis shows that the continuous-action distributions of the studied flow policies are well approximated by Gaussians.
    We propose the Gaussian Action Readout (GAR), which replaces a flow policy's final layer with a Gaussian readout trained on just 0.02\% of eligible states, while keeping the rest of the network frozen.
    It samples with one action-network pass, gives closed-form likelihoods, and achieves near-teacher task success at up to twice the inference speed.
\end{abstract}

\begingroup
\setlength{\intextsep}{0pt}
\setlength{\abovecaptionskip}{4pt}
\begin{figure}[!h]
    \centering
    \includegraphics[width=\linewidth]{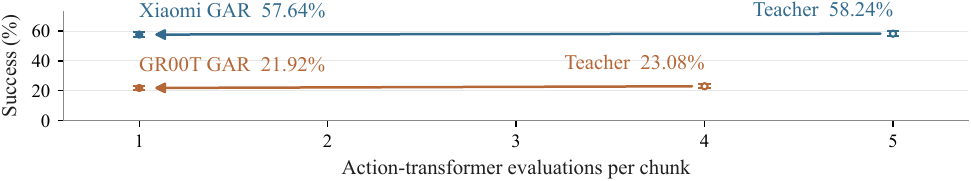}
    \caption{\textbf{Near-teacher success with one action-transformer evaluation.}
        Diagonal GAR uses 6,000 fitting contexts (0.02\% of eligible training states), with teacher networks frozen.
        Each policy is evaluated on 50 RoboCasa365~\citep{robocasa} tasks with 50 episodes per task.
        Bars show pointwise 95\% within-task episode-bootstrap intervals (Appendix~\ref{app:uncertainty}).}
    \label{fig:results_teaser}
\end{figure}
\endgroup

\section{Introduction}

Diffusion and flow-matching policies achieve strong performance in robot manipulation~\citep{chi2023diffusion,pi_0}.
An appealing feature is their ability to model multimodal action distributions, capturing the different valid ways to act in human demonstrations~\citep{chi2023diffusion}.
Generating actions, however, requires iterative sampling with repeated network evaluations, making inference computationally costly.
Their action likelihoods are also unavailable in closed form.
This raises the question: is this distributional flexibility needed for task performance, or could a simpler policy perform just as well while allowing cheaper direct sampling and closed-form likelihoods?

To explore this possibility, we fit separate Gaussians to the translation and rotation action-chunk distributions of Xiaomi-Robotics-1~\citep{xiaomi} and GR00T~\citep{gr00t} at fixed inputs.
We find small approximation errors for both models (Section~\ref{sec:gar_gaussian_approximation}).
These results motivate us to look for a simple, inexpensive way to convert these models into Gaussian policies while retaining their task performance.

A Gaussian is fully specified by its mean and covariance.
We derive relations between these parameters and the flow-matching velocity field (Section~\ref{sec:gar_flow_moments}).
These relations motivate our \emph{Gaussian Action Readout} (GAR), which predicts the Gaussian parameters and binary-command probabilities through small readouts of frozen readout features (Section~\ref{sec:gar_distillation}).
Training only these readouts limits the scope for overfitting, allowing us to fit them using just 6,000 contexts across 300 tasks (0.02\% of eligible training states).

\begingroup
\interlinepenalty=10000
The readouts predict action moments with small median errors (Section~\ref{sec:experiments_moments}).
With GAR, one action-transformer evaluation supplies the parameters for direct Gaussian sampling and closed-form likelihoods.
On 50 RoboCasa365 tasks~\citep{robocasa}, the resulting Gaussian policies achieve similar task performance to Xiaomi and GR00T (Section~\ref{sec:experiments_performance}).
Inference is faster, with a single action-transformer evaluation replacing five flow steps for Xiaomi and four for GR00T.
\par
\endgroup

We also examine how errors in the predicted moments affect task success and whether sampling actions is better than always choosing the mean action (Section~\ref{sec:experiments_diagnostics}).
Sampling can also produce different behaviors over a full execution, even when the action distribution at each decision is Gaussian.
Empirically, we find that our Gaussian policy produces different object-movement orders in successful RoboCasa executions from the same initial context (Section~\ref{sec:experiments_diversity}).

\begin{samepage}
    Our contributions are:
    \begin{itemize}
        \item \textbf{Analysis:} We measure Gaussian approximation errors for Xiaomi and GR00T's translation and rotation distributions across sixteen-action chunks, comparing full and diagonal covariance (Section~\ref{sec:gar_gaussian_approximation}).
        \item \textbf{Method:} Gaussian Action Readout (GAR) predicts action moments and command probabilities through small readouts of frozen readout features.
              One action-transformer evaluation enables direct sampling and closed-form likelihoods (Section~\ref{sec:method}).
        \item \textbf{Experiments:} Gaussian policies retain near-teacher success on 50 RoboCasa365 tasks, with inference speedups of $1.91\times$ for Xiaomi and $1.57\times$ for GR00T (Sections~\ref{sec:experiments_moments}--\ref{sec:experiments_performance}).
              A Gaussian policy can produce diverse behaviors despite its unimodal action distribution (Section~\ref{sec:experiments_diversity}).
    \end{itemize}
\end{samepage}

\section{Related Work}

\paragraph{Flow-based vision-language-action models}
Xiaomi-Robotics-1~\citep{xiaomi} combines a Qwen3-VL backbone~\citep{bai2025qwen3vl} with a diffusion transformer in a Mixture-of-Transformers architecture.
GR00T N1.5~\citep{gr00t,nvidia2025groot15} uses an Eagle 2.5 backbone~\citep{chen2025eagle25} and a diffusion transformer that alternates self-attention over robot-state and action tokens with cross-attention to vision-language features.
Both generate action chunks through iterative flow sampling, requiring repeated action-transformer evaluations.

\paragraph{Simpler action prediction}
OpenVLA-OFT~\citep{kim2025oft} predicts continuous action chunks directly, using an L1 loss against demonstration actions during training.
Its task performance is comparable to that of a diffusion head.
\citet{chen2026simple} improve one-step sampling from flow-matching policies by drawing training inputs more frequently near the pure-noise endpoint.
AdaFlow~\citep{adaflow} uses a learned variance estimate to adapt the number of flow-sampling steps.
\citet{qfat} train a quantization-free transformer (Q-FAT) on demonstrations to predict Gaussian-mixture action distributions with explicit likelihoods.
They report one active density mode at approximately 70\% of rollout steps~\citep{qfat} in Kitchen~\citep{gupta2020relay} and UR3 BlockPush~\citep{kim2022automating}.
\citet{pan2026muchado} argue that multimodality does not fully explain the observed advantage of flow policies and introduce Minimal Iterative Policy (MIP), a lightweight two-step regression policy that approximately matches flow-policy performance.

Direct regression and MIP produce point predictions, while Q-FAT learns Gaussian-mixture action distributions from demonstrations.
We instead ask how close pretrained flow policies are to Gaussian, and whether we can easily convert them into Gaussian policies enabling one-step sampling and closed-form likelihoods.

\paragraph{Policy distillation}
Distillation reduces sampling cost by training a student to reproduce a teacher's outputs.
Consistency Policy~\citep{prasad2024consistency}, One-Step Diffusion Policy~\citep{wang2024onedp} and SnapFlow~\citep{luan2026snapflow} use this approach to generate actions with one or a few network evaluations.
Refined Policy Distillation~\citep{julg2025refined} trains a separate student both to match teacher actions and to maximize task rewards.
Its Proximal Policy Distillation~\citep{spigler2025proximal} baseline fits a Gaussian to ten teacher actions at each input and uses a KL penalty toward this distribution within PPO.
This provides a precedent for Gaussian approximation of a teacher policy, with an input-dependent student mean and a diagonal covariance shared across inputs in the released implementation.

These methods train a complete student action network to imitate the teacher.
GAR reuses the teacher's frozen networks and trains only small readouts to predict a Gaussian action distribution in one pass.

\section{Gaussianity of Pretrained Policies}
\label{sec:gar_gaussian_approximation}

A Gaussian approximation would allow continuous actions to be sampled directly from their mean and covariance.
We first test how well a single Gaussian approximates the continuous-action distributions of pretrained flow policies at fixed observations.

\paragraph{Approximation error.}
At each of 300 fixed observations per teacher, we fit separate Gaussians to translation and rotation commands and evaluate them on held-out teacher samples.
We analyze translation and rotation separately because they have different physical units: combining millimetres and degrees in one distance would require choosing their relative weighting.
Each Gaussian models the three translation or rotation coordinates across all sixteen actions, giving $16\times3=48$ coordinates.
We compare the teacher and Gaussian distributions along many one-dimensional projections.
For each projection, we measure the minimum average displacement needed to match the two distributions.
Averaging these distances gives the sliced 1-Wasserstein distance~\citep{villani2009optimal,bonneel2015sliced}.
Small shifts in command values therefore incur small errors, whereas KL divergence can be large even for small shifts between narrowly concentrated distributions (Appendix~\ref{app:physical_action_scale}).

\begin{figure}[htbp!]
    \centering
    \includegraphics[width=\linewidth]{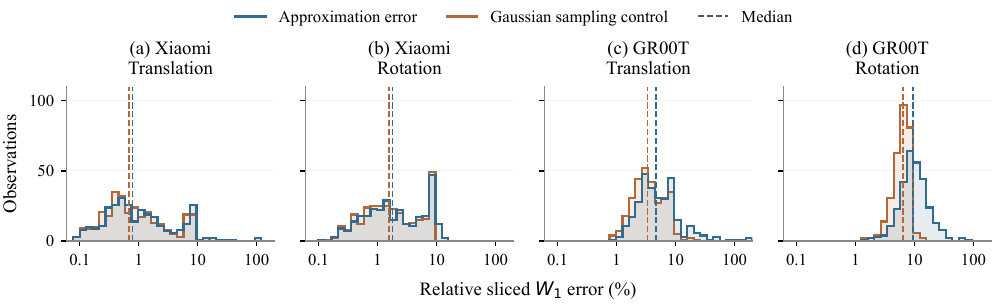}
    \caption{Full-covariance Gaussian approximation across sixteen-action chunks at 300 fixed observations per teacher.
        Histograms show relative sliced 1-Wasserstein errors (\%) for the 48-dimensional translation and rotation groups, normalized by the teacher's projected command magnitude.
        Gaussian sampling controls measure finite-sample error.
        Dashed lines mark medians; logarithmic horizontal axes retain all observations, and lower errors are better.}
    \label{fig:gar_gaussian_error}
\end{figure}

\paragraph{Finite-sample control.}
Even when the underlying action distribution is Gaussian, fitting and evaluating a Gaussian on finite samples can give a nonzero approximation error.
We estimate this baseline error by drawing samples from each fitted Gaussian, fitting a new Gaussian to one subset, and calculating its distance from held-out samples.
Figure~\ref{fig:gar_gaussian_error} compares the approximation errors with this baseline.

With full covariance, Xiaomi's median approximation errors are close to the corresponding controls, whereas GR00T's are somewhat larger (Figure~\ref{fig:gar_gaussian_error}).
Diagonal covariance gives larger median approximation errors for both models (Table~\ref{tab:gar_gaussian_error} and Figure~\ref{fig:gar_gaussian_error_diagonal} in Appendix~\ref{app:chunk-physical-action-scale}).
These results suggest that the continuous-action distributions of these flow policies are already close to Gaussian.

\section{Gaussian Distillation}
\label{sec:method}

We study a simple method for distilling a pretrained flow policy (the teacher) into a Gaussian policy.
Our Gaussian Action Readout (GAR) predicts the Gaussian mean and covariance from the teacher's frozen readout features.

Let $c$ denote the policy context: the task instruction and the image and robot-state observations supplied to the policy, including any observation history.
Let $a=(m,b)$ denote an action chunk, with continuous coordinates $m$ and binary commands $b$.
Let $\pi_T^m(\cdot\mid c)$ denote the teacher's continuous-action marginal, with mean $\mu(c)$ and covariance $\Sigma(c)$.
We approximate this distribution by
\begin{equation}
    \pi_T^m(m\mid c)\approx\mathcal{N}\!\left(m;\mu(c),\Sigma(c)\right).
    \label{eq:gar_teacher_marginal}
\end{equation}
\subsection{Moment Information in Flow Fields}
\label{sec:gar_flow_moments}

A Gaussian is determined by its mean and covariance, so we ask what the flow-matching velocity field tells us about these parameters.

\paragraph{Mean.}
Fix $c$ and draw $m\sim\pi_T^m(\cdot\mid c)$ independently of $\epsilon\sim\mathcal{N}(0,I)$.
The teacher policies are trained with flow matching using a linear interpolation from noise to actions~\citep{lipman2023flow}.
The interpolation and its associated velocity field are
\begin{equation}
    m_t=(1-t)\epsilon+tm,
    \qquad
    v(m_t,t\mid c)=\mathbb{E}[m-\epsilon\mid m_t,c].
    \label{eq:gar_optimal_velocity}
\end{equation}
At $t=0$, the input is independent of the target, giving
\begin{equation}
    \epsilon+v(\epsilon,0\mid c)=\mathbb{E}[m\mid c]=\mu(c).
    \label{eq:gar_initial_mean}
\end{equation}
Thus, a one-unit Euler step recovers the mean, without a Gaussian assumption~\citep[see also][]{mei2026potential}.
When distinct modes represent incompatible actions, their mean can be unsuitable for execution.
For a Gaussian distribution, the mean coincides with the mode.

\paragraph{Diagonal variance.}
Let $i$ index the continuous coordinates of the action chunk.
For a Gaussian target with diagonal covariance $\Sigma(c)=\operatorname{diag}(\sigma_i^2(c))$, $m-\epsilon$ and $m_t$ are jointly Gaussian.
Evaluating the conditional expectation in Equation~\eqref{eq:gar_optimal_velocity} gives (Appendix~\ref{app:gaussian-fm})
\begin{equation}
    \begin{aligned}
        v_{i}(m_t,t\mid c)
         & =\mu_i(c)+\alpha_i(t,c)\bigl(m_{t,i}-t\mu_i(c)\bigr),   \\
        \alpha_i(t,c)
         & =\frac{t\sigma_i^2(c)-(1-t)}{(1-t)^2+t^2\sigma_i^2(c)}.
    \end{aligned}
    \label{eq:gar_diagonal_field}
\end{equation}
For $0<t<1$, a known mean and one velocity query determine $\alpha_i$ whenever $m_{t,i}\neq t\mu_i(c)$, and hence
\begin{equation}
    \sigma_i^2(c)
    =\frac{(1-t)\left[1+(1-t)\alpha_i(t,c)\right]}
    {t\left[1-t\alpha_i(t,c)\right]}.
    \label{eq:gar_diagonal_recovery}
\end{equation}
At $t=0$, $\alpha_i=-1$, so variance is absent from the velocity.

\subsection{Gaussian Action Readouts}
\label{sec:gar_distillation}

\begin{figure}[!htbp]
    \centering
    \includegraphics[width=\linewidth]{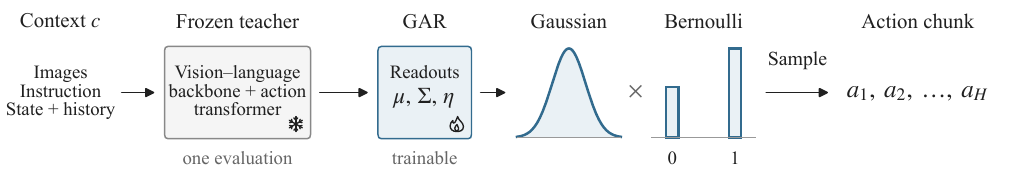}
    \caption{GAR predicts Gaussian parameters and Bernoulli probabilities from one evaluation of the frozen action transformer.
        Only small readouts are trained; sampling continuous actions and binary commands produces an action chunk.}
    \label{fig:gar}
\end{figure}

At $t=0$, the velocity determines the mean but contains no information about the variance.
We ask whether the teacher's internal features still contain enough information to predict both from a single evaluation at $t=0$.
Let $u(c,x,t)$ denote the teacher's \emph{readout features}: the frozen features immediately before its final velocity projection, where $x$ is the noisy action input.
For a fixed noise input $\epsilon_\star$, GAR predicts the distribution parameters from $u(c,\epsilon_\star,0)$ (Figure~\ref{fig:gar}).

\paragraph{Predicting the mean.}
The teacher reads out velocity through a linear projection,
\begin{equation}
    v(x,t\mid c)=W_{\! v}u(c,x,t)+b_v.
    \label{eq:gar_velocity_readout}
\end{equation}
Combining this projection with the mean identity in Equation~\eqref{eq:gar_initial_mean} gives
\begin{equation}
    \begin{aligned}
        \mu(c)
         & =\epsilon_\star+v(\epsilon_\star,0\mid c)            \\
         & =W_{\! v}u(c,\epsilon_\star,0)+(b_v+\epsilon_\star).
    \end{aligned}
    \label{eq:gar_affine_mean_readout}
\end{equation}
At fixed $\epsilon_\star$, the mean is thus an affine function of the features.
This motivates fitting an affine readout of $u(c,\epsilon_\star,0)$ to the mean of teacher samples.

\paragraph{Predicting the variance.}
Under the diagonal Gaussian model, the mean and an intermediate-time velocity query determine each variance through
\begin{equation}
    \begin{aligned}
        \alpha_i(t,c)
         & =\frac{[W_{\! v}u(c,x,t)+b_v]_i-\mu_i(c)}{x_i-t\mu_i(c)}, \\
        \sigma_i^2(c)
         & =\frac{(1-t)\left[1+(1-t)\alpha_i(t,c)\right]}
        {t\left[1-t\alpha_i(t,c)\right]},
    \end{aligned}
    \label{eq:gar_feature_variance_recovery}
\end{equation}
for $0<t<1$ and $x_i\neq t\mu_i(c)$.
These relations connect variance to the intermediate-time features once the mean and query input are known.
Although variance is absent from the time-zero velocity, we ask whether it can still be recovered from the time-zero features.
In both teachers, these features are nearly parallel to those at an intermediate time $s\in(0,1)$, for example $s=0.1$, when $c$ and $x$ are fixed (Appendix~\ref{app:feature-time}).
This motivates learning a variance readout from the time-zero features:
\begin{equation}
    \sigma_\theta^2(c)=g_\theta\!\left(u(c,\epsilon_\star,0)\right),
    \qquad
    \Sigma_\theta(c)=\operatorname{diag}\!\left(\sigma_\theta^2(c)\right),
    \label{eq:gar_variance_readout}
\end{equation}
where $g_\theta$ is an MLP with one hidden layer of 128 units and positive outputs.
We evaluate the accuracy of the predicted moments in Section~\ref{sec:experiments_moments}.
In Section~\ref{sec:experiments_performance}, we compare the task performance of GAR policies using diagonal and full covariance with that of their teachers.

\paragraph{Predicting binary commands.}
Each binary command has two possible outcomes, but its Bernoulli distribution is fully specified by one parameter: its mean $p_j(c)=\Pr(b_j=1\mid c)$.
We represent this distribution by thresholding a Gaussian latent at zero, a construction that also allows correlated latents (Appendix~\ref{app:gaussian-bernoulli}).
In GAR, we use independent Gaussian latents with variance fixed to one.
For each command $j$, an affine readout of the same readout features predicts the latent mean $\eta_{\theta,j}(c)$, giving activation probability $p_{\theta,j}(c)=\Phi(\eta_{\theta,j}(c))$, where $\Phi$ is the standard normal cumulative distribution function.

\paragraph{Fitting.}
The identities above motivate predicting action moments through small readouts of the teacher's readout features.
The deployed teachers generate samples with only five Euler steps for Xiaomi and four for GR00T.
Such coarse integration can change the endpoint distribution and its moments.
The field identities therefore do not guarantee exact agreement with moments estimated from these samples.

At each fitting context, we estimate means, covariances and command frequencies from the deployed teacher's samples.
These estimates provide GAR's fitting targets.
Training only the readouts limits the scope for overfitting the small fitting set.
We fit the mean by least squares, the covariance by normalized squared error, and command probabilities by binary cross-entropy.
Covariance targets are centered on the empirical teacher mean, so mean-prediction errors do not inflate the target covariance.
All fitting uses cached features and teacher statistics (Appendix~\ref{app:moment-readouts}).

\paragraph{Execution.}
At deployment, one action-transformer evaluation supplies the readouts, and we independently sample
\begin{equation}
    \begin{aligned}
        m   & \sim\mathcal{N}\!\left(\mu_\theta(c),\Sigma_\theta(c)\right), \\
        z_j & \sim\mathcal{N}\!\left(\eta_{\theta,j}(c),1\right),
        \qquad b_j=\mathbf{1}\{z_j\geq0\}.
    \end{aligned}
    \label{eq:gar_joint_latent_gaussian}
\end{equation}
The resulting action chunk $a=(m,b)$ requires no velocity-field integration.

\paragraph{Likelihood.}
Unless otherwise stated, GAR uses diagonal covariance for continuous actions, motivated by the argument above.
With binary commands independent of one another and of continuous actions given $c$, the model has the following closed-form likelihood:
\begin{equation}
    p_\theta(m,b\mid c)
    =\mathcal{N}\!\left(m;\mu_\theta(c),\Sigma_\theta(c)\right)
    \prod_j p_{\theta,j}(c)^{b_j}
    \left[1-p_{\theta,j}(c)\right]^{1-b_j},
    \label{eq:gar_likelihood}
\end{equation}
where $p_{\theta,j}(c)=\Phi(\eta_{\theta,j}(c))$ and $\Phi$ is the standard normal cumulative distribution function.
Once the parameters have been computed, this expression scores candidate action chunks without further network evaluations or flow integration.

\section{Experiments}
\label{sec:experiments}

We evaluate moment accuracy and task performance (Sections~\ref{sec:experiments_moments}--\ref{sec:experiments_performance}), then isolate the roles of Gaussian approximation, moment prediction and sampling (Section~\ref{sec:experiments_diagnostics}).
Repeated executions test whether Gaussian actions can produce different successful behaviors (Section~\ref{sec:experiments_diversity}).

Our main experiments use RoboCasa365, a simulation benchmark with 365 household manipulation tasks across 2,500 kitchen environments~\citep{robocasa}.
Its tasks range from basic skills, such as opening drawers, to multistep activities, such as preparing coffee.
The teachers, Xiaomi-Robotics-1 and GR00T N1.5, were trained on Human300 demonstrations covering 300 of these tasks.
Appendix~\ref{app:setting} includes examples of successful GAR executions.

Unless stated otherwise, we fit the readouts on 6,000 contexts, taking four evenly spaced interior states from each of five randomly selected human demonstrations per task (Appendix~\ref{app:context-sampling}).
Control evaluation then compares the policies on the 50-task target suite, which includes seen tasks and unseen composite tasks, with 50 episodes per policy per task.

We also evaluate GAR with the \mbox{Xiaomi-Robotics-0} teacher~\citep{cai2026xiaomi} on the ten LIBERO-Long tasks~\citep{liu2023libero}.
These are multi-step manipulation tasks, such as placing a bowl in a drawer and closing it.

\begingroup
\setlength{\intextsep}{6pt}
\subsection{Moment Recovery}
\label{sec:experiments_moments}

We first test how accurately small readouts predict the teacher's mean and marginal standard deviations (SDs).
With 6,000 fitting contexts, median errors in the predicted mean and SDs are below 10\% of the norm of the teacher's mean action chunk for both teachers (Table~\ref{tab:gar_moment_fit}).
For GR00T, median variance predictions are close to the teacher estimates, but individual predictions still show substantial scatter.
For Xiaomi, median predictions broadly track the teacher estimates, but the largest variances are systematically underestimated.
Individual predictions also show greater scatter (Figure~\ref{fig:gar-variance-calibration}).

\begin{table}[!htbp]
    \small
    \setlength{\tabcolsep}{5pt}
    \caption{Moment prediction on 600 held-out contexts per teacher, using estimates from teacher samples as references.
        Entries give median [25th, 75th percentile] errors, with lower values better.
        Mean and SD errors are $100\|\mu_\theta(c)-\widehat\mu_c\|_2/\|\widehat\mu_c\|_2$ and $100\|\sigma_\theta(c)-\widehat\sigma_c\|_2/\|\widehat\mu_c\|_2$, respectively, where $\sigma$ denotes the vector of marginal SDs.
        Diagonal and full variants share the mean readout.
        Errors use normalized coordinates for Xiaomi and decoded coordinates for GR00T, so comparisons are within each teacher.
        More fitting contexts reduce median diagonal-SD error, with little change in mean error.}
    \label{tab:gar_moment_fit}
    \newcommand{\momenterror}[3]{%
        \makebox[2em][r]{#1}\,[\makebox[2em][r]{#2},\,\makebox[2.5em][r]{#3}]}
    \centering
        \begin{tabular}{ll*{3}{>{\centering\arraybackslash}p{8.5em}}}
            \toprule
            Teacher & Fitting contexts                  & Mean error (\%)
                    & \multicolumn{2}{c}{SD error (\%)}                                                                                                       \\
            \cmidrule(lr){4-5}
                    &                                   &                                 & Diagonal                        & Full                            \\
            \midrule
            Xiaomi  & 6,000                             & \momenterror{4.63}{3.09}{7.73}  & \momenterror{5.58}{3.61}{9.03}  & \momenterror{2.91}{1.72}{5.95}  \\
            Xiaomi  & 24,000                            & \momenterror{4.54}{3.00}{7.37}  & \momenterror{4.35}{2.69}{7.37}  & \momenterror{2.92}{1.56}{5.57}  \\
            GR00T   & 6,000                             & \momenterror{8.43}{5.98}{12.91} & \momenterror{9.76}{6.51}{16.42} & \momenterror{8.09}{5.36}{13.42} \\
            GR00T   & 24,000                            & \momenterror{8.29}{5.86}{12.55} & \momenterror{8.47}{5.74}{14.65} & \momenterror{7.61}{5.08}{12.21} \\
            \bottomrule
        \end{tabular}
\end{table}

\paragraph{Data requirements.}
To test whether more supervision improves moment recovery, we increase the fitting set from 6,000 to 24,000 contexts using additional demonstrations.
Median diagonal SD error drops by 1.23 percentage points for Xiaomi and 1.29 for GR00T, while median mean error changes by less than 0.2 percentage points.

\paragraph{Fitting cost.}
Once features and targets are cached, fitting Xiaomi's 6,000-context diagonal readouts takes approximately 115 seconds, including the joint fit that supplies the command readout.
Teacher sampling, feature extraction and untimed setup are additional costs (Appendix~\ref{app:fitting-cost}).

\clearpage
\subsection{Task Performance}
\label{sec:experiments_performance}

We next test whether GAR retains the teachers' task success with one action-transformer evaluation.
Across the 50-task RoboCasa365 suite, GAR achieves overall success rates close to those of both teachers (Table~\ref{tab:gar_success}).

\begin{table}[!htbp]
    \small
    \setlength{\tabcolsep}{3pt}
    \caption{RoboCasa365 success (\%) across 50 tasks and 2,500 unpaired episodes per policy; higher is better.
        Groups contain 18 seen atomic, 16 seen composite and 16 unseen composite tasks, with 50 episodes per task.
        Raw one-step uses the teacher without fitted readouts.
        Mean-only GAR retains stochastic binary commands; Gaussian variants use predicted variance.
        Context counts refer to mean and covariance fitting; command readouts use 6,000 contexts.
        Execution, fitting and uncertainty are detailed in Appendices~\ref{app:setting},~\ref{app:readout-fitting} and~\ref{app:uncertainty}, respectively.}
    \label{tab:gar_success}
    \label{tab:gar_task_groups}
    \centering
        \begin{tabular}{lllrrrrr}
            \toprule
            Teacher & \shortstack{Fitting                                                                                                         \\contexts} & Task group & Teacher & \shortstack{Raw\\one-step} & Mean only & Diagonal & Full  \\
            \midrule
            Xiaomi  & 6,000               & \textbf{Overall} & \textbf{58.24} & \textbf{55.76} & \textbf{56.64} & \textbf{57.64} & \textbf{57.28} \\
                    &                     & Atomic seen      & 81.11          & 81.22          & 78.44          & 80.56          & 81.67          \\
                    &                     & Composite seen   & 58.75          & 52.63          & 58.13          & 58.00          & 57.38          \\
                    &                     & Composite unseen & 32.00          & 30.25          & 30.63          & 31.50          & 29.75          \\
            \addlinespace
                    & 24,000              & \textbf{Overall} & \textbf{58.24} & \textbf{55.76} & \textbf{56.44} & \textbf{57.72} & \textbf{56.44} \\
                    &                     & Atomic seen      & 81.11          & 81.22          & 80.11          & 81.00          & 81.11          \\
                    &                     & Composite seen   & 58.75          & 52.63          & 54.50          & 56.75          & 55.25          \\
                    &                     & Composite unseen & 32.00          & 30.25          & 31.75          & 32.50          & 29.88          \\
            \midrule
            GR00T   & 6,000               & \textbf{Overall} & \textbf{23.08} & \textbf{19.28} & \textbf{19.72} & \textbf{21.92} & \textbf{20.32} \\
                    &                     & Atomic seen      & 46.00          & 43.00          & 44.33          & 45.78          & 43.89          \\
                    &                     & Composite seen   & 15.38          & 8.13           & 7.75           & 11.75          & 11.50          \\
                    &                     & Composite unseen & 5.00           & 3.75           & 4.00           & 5.25           & 2.63           \\
            \addlinespace
                    & 24,000              & \textbf{Overall} & \textbf{23.08} & \textbf{19.28} & \textbf{20.20} & \textbf{21.44} & \textbf{21.16} \\
                    &                     & Atomic seen      & 46.00          & 43.00          & 43.22          & 44.11          & 44.67          \\
                    &                     & Composite seen   & 15.38          & 8.13           & 10.50          & 12.63          & 12.63          \\
                    &                     & Composite unseen & 5.00           & 3.75           & 4.00           & 4.75           & 3.25           \\
            \bottomrule
        \end{tabular}
\end{table}

The benefit of fitting GAR is clearest on seen composite tasks.
Mean-only GAR nearly closes Xiaomi's gap to its teacher.
For GR00T, retaining continuous-action variance improves success over mean-only execution, although a gap to the teacher remains.
GR00T's small gap on unseen composite tasks is less informative because teacher success is only 5\%.

\paragraph{Full covariance and fitting data.}
Full covariance does not improve success over diagonal sampling, despite lower marginal-SD error, which omits correlation errors.
More fitting contexts improve diagonal SD predictions without consistently improving task success.

\paragraph{Inference cost.}
GAR replaces five action-transformer evaluations for Xiaomi and four for GR00T with one.
For the 6,000-context diagonal policies, batch-one latency falls from 220.25 to 115.36\,ms for Xiaomi ($1.91\times$) and from 68.01 to 43.30\,ms for GR00T ($1.57\times$).
These timings include vision-language processing, which limits the total speedup (Appendix~\ref{app:gar-inference-cost}).

\paragraph{LIBERO-Long.}
With the \mbox{Xiaomi-Robotics-0} teacher, 6,000-context diagonal GAR achieves 95.4\% success across all ten LIBERO-Long tasks, compared with 96.6\% for the teacher and 96.0\% for raw one-step execution.
This extends near-teacher Gaussian control to a second benchmark.
Appendices~\ref{app:libero-setting},~\ref{app:libero-fitting} and~\ref{app:libero-long} give the protocol, fitting procedure and per-task results.

\endgroup

\clearpage
\subsection{Gaussian Approximation and Sampling}
\label{sec:experiments_diagnostics}

We ask whether shape errors in the Gaussian approximation affect task success, separately from errors in GAR's predicted mean and covariance.
We also examine why executing the mean can work well and whether sampling improves on it.
The controls use the sixteen seen composite tasks, where GR00T's teacher gap is largest (Table~\ref{tab:gar_success}).

\paragraph{Mean-only execution.}
The Gaussian approximation offers an explanation for strong mean-only performance: for a Gaussian, the mean is also the most likely action.
We examine two ways of estimating this mean: raw one-step execution, motivated by Equation~\eqref{eq:gar_initial_mean}, and mean-only GAR, fitted to the mean of the deployed teacher's finite-step samples.
Both retain much of the teachers' success (Table~\ref{tab:gar_success}), consistent with this explanation.

\paragraph{Gaussian approximation.}
A Gaussian action policy could lose performance because its distribution has the wrong shape or because its mean and covariance are poorly estimated.
To test the shape assumption separately, we set GAR's moment predictions aside.
At each observation, we draw teacher samples and fit a Gaussian to their mean and full covariance.
We then compare sampling from this Gaussian with sampling from the teacher, keeping GAR's binary-command rule fixed in both policies.
Gaussian sampling achieves similar success for GR00T and slightly higher success for Xiaomi (Table~\ref{tab:gar_diagnostics}), showing no clear performance cost from the Gaussian approximation in these controls.

\paragraph{Mean estimation.}
We next ask whether errors in GAR's predicted mean reduce task success.
We compare two policies that execute a mean action without motion noise: one uses GAR's predicted mean, while the other uses the average of teacher samples at the current observation.
Both retain the same GAR command rule, so the comparison changes only how the mean is estimated.
The teacher-estimated mean improves GR00T's success, with no clear change for Xiaomi (Table~\ref{tab:gar_diagnostics}).
This identifies mean estimation as a limitation of GR00T's mean-only control.

\begin{table}[!htbp]
    \small
    \setlength{\tabcolsep}{3pt}
    \caption{Seen-composite success (\%) under Table~\ref{tab:gar_success}'s protocol (Appendix~\ref{app:gr00t-recovery}): 32 episodes per task and model for each intervention, 50 for reused references.
        Differences are second minus first (pp), with 95\% independent within-task bootstrap intervals.
        All policies retain GAR commands except raw one-step, which uses native decoding.}
    \label{tab:gar_diagnostics}
    \centering
    \begin{tabularx}{\linewidth}{@{}Xrrrr@{}}
        \toprule
        & \multicolumn{2}{c}{GR00T} & \multicolumn{2}{c}{Xiaomi} \\
        \cmidrule(lr){2-3}\cmidrule(l){4-5}
        Continuous-motion policy & Success & Difference & Success & Difference \\
        \midrule
        \multicolumn{5}{@{}l}{\textit{Benchmark references from Table~\ref{tab:gar_success}: 50 episodes per task}} \\
        Raw one-step
            & 8.13 & --- & 52.63 & --- \\
        Mean-only GAR, 6,000 contexts
            & 7.75 & --- & 58.13 & --- \\
        \midrule
        \multicolumn{5}{@{}l}{\textit{Distribution shape: Gaussian versus teacher sampling}} \\
        Full Gaussian, teacher moments
            & 11.72 & --- & 59.38 & --- \\
        Teacher sampling, GAR commands
            & 11.91 & $+0.20\ [-3.32,3.71]$ & 57.81 & $-1.56\ [-7.03,3.71]$ \\
        \addlinespace
        \multicolumn{5}{@{}l}{\textit{Moment prediction: GAR versus teacher-estimated moments}} \\
        Diagonal Gaussian, GAR moments
            & 11.75 & --- & 58.00 & --- \\
        Diagonal Gaussian, teacher moments
            & 13.09 & $+1.34\ [-2.02,4.93]$ & 58.59 & $+0.59\ [-4.33,5.25]$ \\
        \midrule
        \multicolumn{5}{@{}l}{\textit{Sampling around a teacher-estimated mean}} \\
        No motion noise
            & 11.33 & --- & 55.27 & --- \\
        Gaussian noise with GAR variances
            & 14.65 & $+3.32\ [-0.39,7.23]$ & 60.16 & $+4.88\ [-0.20,10.16]$ \\
        \bottomrule
    \end{tabularx}
\end{table}

\paragraph{Sampling.}
Sampling improves overall success over mean-only GAR in the main benchmark (Table~\ref{tab:gar_success}).
We ask whether this benefit remains when the mean is estimated directly from the teacher.
Keeping that mean estimator and GAR's command rule fixed, we compare executing the mean with sampling around it using GAR's predicted variances.
Sampling gives higher observed success for both models (Table~\ref{tab:gar_diagnostics}), although both confidence intervals include zero.
The results suggest that sampling can remain useful even after replacing GAR's mean prediction.

\clearpage
\subsection{Multimodal behavior}
\label{sec:experiments_diversity}

Beyond improving task completion, Gaussian sampling can produce different successful behaviors.
A single Gaussian at each decision can yield multimodal behavior through diffusion in physical time.

\paragraph{Diffusion in physical time.}
For direct position updates, Gaussian action sampling gives
\begin{equation}
    s_{k+1}=s_k+\mu(s_k)+\Sigma^{1/2}\xi_k,
    \qquad \xi_k\overset{\mathrm{iid}}{\sim}\mathcal N(0,I).
    \label{eq:gaussian-position-update}
\end{equation}
This update is a discretized diffusion in physical time (Appendix~\ref{app:toy-gaussian}).

Each transition is Gaussian conditional on the current state, but the distribution of later states can be multimodal.
A nonlinear mean $\mu$ can amplify small random deviations into distinct routes.
In our constructed example, trajectories from the same start separate around an obstacle before returning toward a common goal (Figure~\ref{fig:gaussian-behavior}).

\begingroup
\setlength{\intextsep}{6pt}
\begin{figure}[!ht]
    \centering
    \includegraphics[width=0.58\linewidth,trim=0 18 0 18,clip]{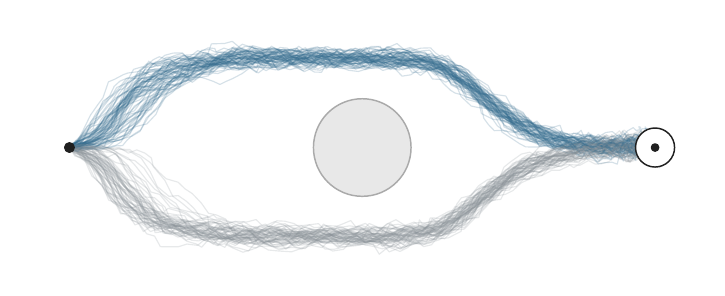}
    \caption{
        Gaussian actions produce two routes in a constructed example.
        All 128 rollouts start at the left dot, pass on either side of the obstacle, and approach the right circle.
        Appendix~\ref{app:toy-gaussian} specifies the state-dependent mean and constant covariance.
    }
    \label{fig:gaussian-behavior}
\end{figure}
\endgroup

\FloatBarrier
\paragraph{Different successful movement orders.}
Xiaomi's Diagonal policy also produces distinct successful movement orders from the same initial context.
In PackIdenticalLunches, the robot must place one garlic and one chicken drumstick into each container.
Three of 32 executions succeed, with two different object-movement orders among these successes (Figure~\ref{fig:robocasa-successful-orders}; Appendix~\ref{app:successful-orders}).

\begingroup
\setlength{\intextsep}{6pt}
\begin{figure}[!ht]
    \centering
    \includegraphics[width=\linewidth,trim=0 4 0 7,clip]{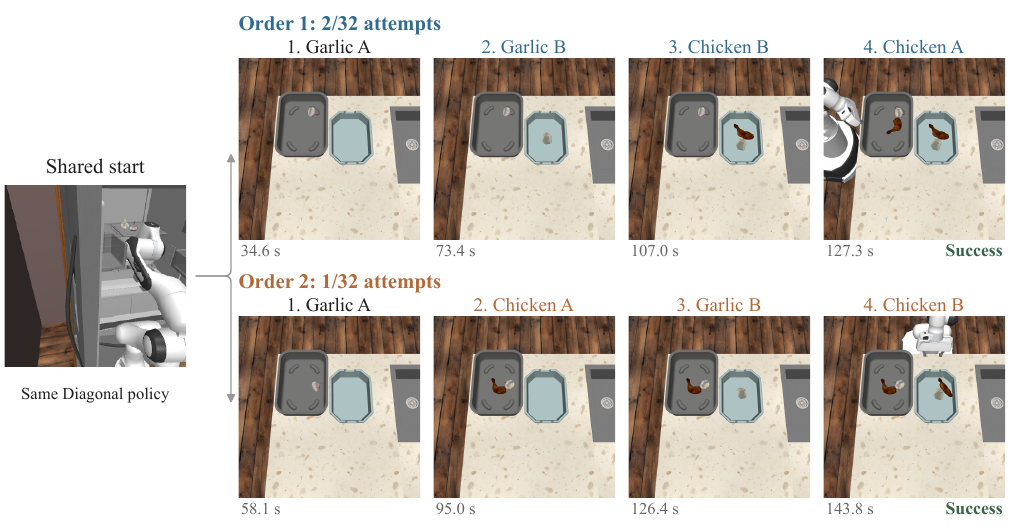}
    \caption{
        Xiaomi's diagonal GAR produces two successful movement orders from the same start.
        The task is to pack one garlic and one chicken drumstick into each container; A and B distinguish objects of each type.
        Left: shared initial policy-camera view.
        Right: recorded simulator states after successive transfers, rendered from a fixed external camera and labeled with elapsed physical time.
        The two orders account for all three successes among 32 attempts: two follow the upper order and one follows the lower order (Appendix~\ref{app:successful-orders}).
    }
    \label{fig:robocasa-successful-orders}
\end{figure}
\endgroup

\FloatBarrier
\section{Conclusion}

Gaussian Action Readouts fitted on 0.02\% of eligible training states enable direct sampling and closed-form likelihoods from one frozen action-transformer evaluation.
GAR retains near-teacher success for Xiaomi and GR00T while reducing inference latency.
Neither model shows a clear success advantage for empirical teacher motion over a full Gaussian fitted to its moments.
Sampling gives higher observed success than mean-only execution, including with teacher-estimated means, although the latter gains remain uncertain.
Through state feedback, Gaussian actions can also produce distinct successful behaviors from the same initial context.

\section*{Reproducibility Statement}

Appendix~\ref{app:setting} specifies the checkpoints and evaluation protocols.
Appendix~\ref{app:readout-fitting} gives data splits, teacher sampling and readout fitting.
Appendices~\ref{app:physical_action_scale} and~\ref{app:moment-recovery} describe the Gaussian-approximation and moment-prediction measurements.
Appendices~\ref{app:performance} and~\ref{app:diversity} give the task-performance and diversity protocols, including uncertainty calculations.
Code will be released soon.

\section*{AI Use Statement}

We used generative AI tools to assist with hypothesis development, experiment design, mathematical derivations, code, analysis, and manuscript preparation.
The authors directed the research and take responsibility for the final text, claims, and results.

\clearpage
\bibliographystyle{iclr2027_conference}
\bibliography{references}

\begin{thebibliography}{31}
\providecommand{\natexlab}[1]{#1}
\providecommand{\url}[1]{\texttt{#1}}
\expandafter\ifx\csname urlstyle\endcsname\relax
  \providecommand{\doi}[1]{doi: #1}\else
  \providecommand{\doi}{doi: \begingroup \urlstyle{rm}\Url}\fi

\bibitem[Bai et~al.(2025)Bai, Cai, Chen, Chen, Chen, Cheng, Deng, Ding, Gao,
  Ge, et~al.]{bai2025qwen3vl}
Shuai Bai, Yuxuan Cai, Ruizhe Chen, Keqin Chen, Xionghui Chen, Zesen Cheng,
  Lianghao Deng, Wei Ding, Chang Gao, Chunjiang Ge, et~al.
\newblock {Qwen3-VL} technical report.
\newblock \emph{arXiv preprint arXiv:2511.21631}, 2025.
\newblock URL \url{https://arxiv.org/abs/2511.21631}.

\bibitem[Black et~al.(2025)Black, Brown, Driess, Esmail, Equi, Finn, Fusai,
  Groom, Hausman, Ichter, et~al.]{pi_0}
Kevin Black, Noah Brown, Danny Driess, Adnan Esmail, Michael~Robert Equi,
  Chelsea Finn, Niccolo Fusai, Lachy Groom, Karol Hausman, Brian Ichter, et~al.
\newblock {$\pi_0$}: A vision-language-action flow model for general robot
  control.
\newblock In \emph{Proceedings of Robotics: Science and Systems}, 2025.
\newblock \doi{10.15607/RSS.2025.XXI.010}.
\newblock URL \url{https://roboticsproceedings.org/rss21/p010.html}.

\bibitem[Bonneel et~al.(2015)Bonneel, Rabin, Peyr{\'e}, and
  Pfister]{bonneel2015sliced}
Nicolas Bonneel, Julien Rabin, Gabriel Peyr{\'e}, and Hanspeter Pfister.
\newblock Sliced and radon {Wasserstein} barycenters of measures.
\newblock \emph{Journal of Mathematical Imaging and Vision}, 51\penalty0
  (1):\penalty0 22--45, 2015.
\newblock \doi{10.1007/s10851-014-0506-3}.
\newblock URL \url{https://doi.org/10.1007/s10851-014-0506-3}.

\bibitem[Cai et~al.(2026)Cai, Guo, He, Jin, Li, Lin, Liu, Liu, Ma, Ma, Qiu, Qu,
  Su, Sun, Wang, Wang, Wang, Wu, Xiang, Yang, Ye, Zhang, and
  Zhou]{cai2026xiaomi}
Rui Cai, Jun Guo, Xinze He, Piaopiao Jin, Jie Li, Bingxuan Lin, Futeng Liu, Wei
  Liu, Fei Ma, Kun Ma, Feng Qiu, Heng Qu, Yifei Su, Qiao Sun, Dong Wang,
  Donghao Wang, Yunhong Wang, Rujie Wu, Diyun Xiang, Yu~Yang, Hangjun Ye, Yuan
  Zhang, and Quanyun Zhou.
\newblock {\mbox{Xiaomi-Robotics-0}: An Open-Sourced Vision-Language-Action
  Model with Real-Time Execution}.
\newblock \emph{arXiv preprint arXiv:2602.12684}, 2026.
\newblock URL \url{https://arxiv.org/abs/2602.12684}.

\bibitem[Chen et~al.(2025)Chen, Li, Wang, Jiang, Liu, Lu, Huang, Byeon, Le,
  Ehrlich, Lu, Wang, Catanzaro, Kautz, Tao, Yu, and Liu]{chen2025eagle25}
Guo Chen, Zhiqi Li, Shihao Wang, Jindong Jiang, Yicheng Liu, Lidong Lu, De-An
  Huang, Wonmin Byeon, Matthieu Le, Max Ehrlich, Tong Lu, Limin Wang, Bryan
  Catanzaro, Jan Kautz, Andrew Tao, Zhiding Yu, and Guilin Liu.
\newblock {Eagle 2.5}: Boosting long-context post-training for frontier
  vision-language models.
\newblock In \emph{Advances in Neural Information Processing Systems},
  volume~38, pp.\  91077--91100, 2025.
\newblock \doi{10.52202/085713-3047}.
\newblock URL
  \url{https://proceedings.neurips.cc/paper_files/paper/2025/hash/833295cb9278a3ba973842a94ea68e3c-Abstract-Conference.html}.

\bibitem[Chen et~al.(2026)Chen, Zhang, Gong, and Qiu]{chen2026simple}
Yitong Chen, Shiduo Zhang, Jingjing Gong, and Xipeng Qiu.
\newblock Let it be simple: One-step action generation for
  vision-language-action models.
\newblock \emph{arXiv preprint arXiv:2606.05737}, 2026.
\newblock URL \url{https://arxiv.org/abs/2606.05737}.

\bibitem[Chi et~al.(2023)Chi, Feng, Du, Xu, Cousineau, Burchfiel, and
  Song]{chi2023diffusion}
Cheng Chi, Siyuan Feng, Yilun Du, Zhenjia Xu, Eric Cousineau, Benjamin~CM
  Burchfiel, and Shuran Song.
\newblock Diffusion policy: Visuomotor policy learning via action diffusion.
\newblock In \emph{Proceedings of Robotics: Science and Systems}, 2023.
\newblock \doi{10.15607/RSS.2023.XIX.026}.
\newblock URL \url{https://roboticsproceedings.org/rss19/p026.html}.

\bibitem[Gupta et~al.(2020)Gupta, Kumar, Lynch, Levine, and
  Hausman]{gupta2020relay}
Abhishek Gupta, Vikash Kumar, Corey Lynch, Sergey Levine, and Karol Hausman.
\newblock Relay policy learning: Solving long-horizon tasks via imitation and
  reinforcement learning.
\newblock In \emph{Proceedings of the Conference on Robot Learning}, volume 100
  of \emph{Proceedings of Machine Learning Research}, pp.\  1025--1037. PMLR,
  2020.
\newblock URL \url{https://proceedings.mlr.press/v100/gupta20a.html}.

\bibitem[Hu et~al.(2024)Hu, Liu, Liu, and Liu]{adaflow}
Xixi Hu, Bo~Liu, Xingchao Liu, and Qiang Liu.
\newblock {AdaFlow}: Imitation learning with variance-adaptive flow-based
  policies.
\newblock In \emph{Advances in Neural Information Processing Systems},
  volume~37, pp.\  138836--138858, 2024.
\newblock \doi{10.52202/079017-4405}.
\newblock URL
  \url{https://proceedings.neurips.cc/paper_files/paper/2024/hash/fa9cfdb49f7de3dee54007f84c0745b1-Abstract-Conference.html}.

\bibitem[J{\"u}lg et~al.(2025)J{\"u}lg, Burgard, and Walter]{julg2025refined}
Tobias J{\"u}lg, Wolfram Burgard, and Florian Walter.
\newblock Refined policy distillation: From {VLA} generalists to {RL} experts.
\newblock In \emph{IEEE/RSJ International Conference on Intelligent Robots and
  Systems}, pp.\  11677--11684, 2025.
\newblock \doi{10.1109/IROS60139.2025.11246761}.
\newblock URL \url{https://doi.org/10.1109/IROS60139.2025.11246761}.

\bibitem[Kim et~al.(2022)Kim, hyeon Park, Cho, and Kim]{kim2022automating}
Jigang Kim, J.~hyeon Park, Daesol Cho, and H.~Jin Kim.
\newblock Automating reinforcement learning with example-based resets.
\newblock \emph{IEEE Robotics and Automation Letters}, 7\penalty0 (3):\penalty0
  6606--6613, 2022.
\newblock \doi{10.1109/LRA.2022.3173039}.
\newblock URL \url{https://doi.org/10.1109/LRA.2022.3173039}.

\bibitem[Kim et~al.(2025)Kim, Finn, and Liang]{kim2025oft}
Moo~Jin Kim, Chelsea Finn, and Percy Liang.
\newblock Fine-tuning vision-language-action models: Optimizing speed and
  success.
\newblock In \emph{Proceedings of Robotics: Science and Systems}, 2025.
\newblock \doi{10.15607/RSS.2025.XXI.017}.
\newblock URL \url{https://roboticsproceedings.org/rss21/p017.html}.

\bibitem[Lipman et~al.(2023)Lipman, Chen, Ben-Hamu, Nickel, and
  Le]{lipman2023flow}
Yaron Lipman, Ricky T.~Q. Chen, Heli Ben-Hamu, Maximilian Nickel, and Matt Le.
\newblock Flow matching for generative modeling.
\newblock In \emph{International Conference on Learning Representations}, 2023.
\newblock URL \url{https://openreview.net/forum?id=PqvMRDCJT9t}.

\bibitem[Liu et~al.(2023)Liu, Zhu, Gao, Feng, Liu, Zhu, and
  Stone]{liu2023libero}
Bo~Liu, Yifeng Zhu, Chongkai Gao, Yihao Feng, Qiang Liu, Yuke Zhu, and Peter
  Stone.
\newblock {LIBERO}: Benchmarking knowledge transfer for lifelong robot
  learning.
\newblock In \emph{Advances in Neural Information Processing Systems},
  volume~36, pp.\  44776--44791, 2023.
\newblock \doi{10.52202/075280-1939}.
\newblock URL
  \url{https://proceedings.neurips.cc/paper_files/paper/2023/hash/8c3c666820ea055a77726d66fc7d447f-Abstract-Datasets_and_Benchmarks.html}.

\bibitem[Loshchilov \& Hutter(2019)Loshchilov and
  Hutter]{loshchilov2019decoupled}
Ilya Loshchilov and Frank Hutter.
\newblock Decoupled weight decay regularization.
\newblock In \emph{International Conference on Learning Representations}, 2019.
\newblock URL \url{https://openreview.net/forum?id=Bkg6RiCqY7}.

\bibitem[Luan et~al.(2026)Luan, Li, Zhao, Zhang, Wu, and Ma]{luan2026snapflow}
Wuyang Luan, Junhui Li, Weiguang Zhao, Wenjian Zhang, Tieru Wu, and Rui Ma.
\newblock {SnapFlow}: One-step action generation for flow-matching {VLAs} via
  progressive self-distillation.
\newblock \emph{arXiv preprint arXiv:2604.05656}, 2026.
\newblock URL \url{https://arxiv.org/abs/2604.05656}.

\bibitem[Matou{\v{s}}ek(1998)]{matousek1998discrepancy}
Ji{\v{r}}{\'i} Matou{\v{s}}ek.
\newblock On the {$L_2$}-discrepancy for anchored boxes.
\newblock \emph{Journal of Complexity}, 14\penalty0 (4):\penalty0 527--556,
  1998.
\newblock \doi{10.1006/jcom.1998.0489}.
\newblock URL \url{https://doi.org/10.1006/jcom.1998.0489}.

\bibitem[Mei et~al.(2026)Mei, He, Cao, Wang, Zhu, Zhou, Wang, Xin, Cheng, Yang,
  Wang, Zheng, Huang, Chen, and Wang]{mei2026potential}
Yunpeng Mei, Jiakai He, Hongjie Cao, Chenyu Wang, Xiaowen Zhu, Yihan Zhou,
  Jiamin Wang, Chenbo Xin, Peng Cheng, Yuxuan Yang, Yijie Wang, Xinhu Zheng,
  Gao Huang, Jie Chen, and Gang Wang.
\newblock Potential-guided flow matching for vision-language-action policy
  improvement.
\newblock \emph{arXiv preprint arXiv:2606.04968}, 2026.
\newblock URL \url{https://arxiv.org/abs/2606.04968}.

\bibitem[Nasiriany et~al.(2026)Nasiriany, Nasiriany, Maddukuri, and
  Zhu]{robocasa}
Soroush Nasiriany, Sep Nasiriany, Abhiram Maddukuri, and Yuke Zhu.
\newblock {RoboCasa365: A Large-Scale Simulation Framework for Training and
  Benchmarking Generalist Robots}.
\newblock In \emph{International Conference on Learning Representations},
  volume 2026, pp.\  98643--98667, 2026.
\newblock URL
  \url{https://proceedings.iclr.cc/paper_files/paper/2026/hash/a05003fdb1e9562ab0c0a9719ea4de10-Abstract-Conference.html}.

\bibitem[{NVIDIA}(2025)]{nvidia2025groot15}
{NVIDIA}.
\newblock {GR00T-N1.5-3B}: Model card, 2025.
\newblock URL \url{https://huggingface.co/nvidia/GR00T-N1.5-3B}.

\bibitem[{NVIDIA} et~al.(2025){NVIDIA}, Bjorck, Casta{\~n}eda, Cherniadev, Da,
  Ding, Fan, Fang, Fox, Hu, Huang, et~al.]{gr00t}
{NVIDIA}, Johan Bjorck, Fernando Casta{\~n}eda, Nikita Cherniadev, Xingye Da,
  Runyu Ding, Linxi~{``Jim''} Fan, Yu~Fang, Dieter Fox, Fengyuan Hu, Spencer
  Huang, et~al.
\newblock {GR00T N1}: An open foundation model for generalist humanoid robots.
\newblock \emph{arXiv preprint arXiv:2503.14734}, 2025.
\newblock URL \url{https://arxiv.org/abs/2503.14734}.

\bibitem[Pan et~al.(2026)Pan, Anantharaman, Huang, Jin, Pfrommer, Yuan,
  Permenter, Qu, Boffi, Shi, and Simchowitz]{pan2026muchado}
Chaoyi Pan, Giridharan Anantharaman, Nai-Chieh Huang, Claire Jin, Daniel
  Pfrommer, Chenyang Yuan, Frank Permenter, Guannan Qu, Nicholas Boffi, Guanya
  Shi, and Max Simchowitz.
\newblock Much ado about noising: Dispelling the myths of generative robotic
  control.
\newblock In \emph{International Conference on Learning Representations},
  volume 2026, pp.\  90575--90614, 2026.
\newblock URL
  \url{https://proceedings.iclr.cc/paper_files/paper/2026/hash/9299ca9ed58731945e934adb5b71728c-Abstract-Conference.html}.

\bibitem[Prasad et~al.(2024)Prasad, Lin, Wu, Zhou, and
  Bohg]{prasad2024consistency}
Aaditya Prasad, Kevin Lin, Jimmy Wu, Linqi Zhou, and Jeannette Bohg.
\newblock Consistency policy: Accelerated visuomotor policies via consistency
  distillation.
\newblock In \emph{Proceedings of Robotics: Science and Systems}, 2024.
\newblock \doi{10.15607/RSS.2024.XX.071}.
\newblock URL \url{https://roboticsproceedings.org/rss20/p071.html}.

\bibitem[Sheebaelhamd et~al.(2025)Sheebaelhamd, Tschannen, Muehlebach, and
  Vernade]{qfat}
Ziyad Sheebaelhamd, Michael Tschannen, Michael Muehlebach, and Claire Vernade.
\newblock Quantization-free autoregressive action transformer.
\newblock In \emph{Advances in Neural Information Processing Systems},
  volume~38, pp.\  81187--81212, 2025.
\newblock \doi{10.52202/085713-2712}.
\newblock URL
  \url{https://proceedings.neurips.cc/paper_files/paper/2025/hash/7499cd7bf0b08d810807c191bc177545-Abstract-Conference.html}.

\bibitem[Sobol'(1967)]{sobol1967distribution}
I.~M. Sobol'.
\newblock On the distribution of points in a cube and the approximate
  evaluation of integrals.
\newblock \emph{USSR Computational Mathematics and Mathematical Physics},
  7\penalty0 (4):\penalty0 86--112, 1967.
\newblock \doi{10.1016/0041-5553(67)90144-9}.
\newblock URL \url{https://doi.org/10.1016/0041-5553(67)90144-9}.

\bibitem[Spigler(2025)]{spigler2025proximal}
Giacomo Spigler.
\newblock Proximal policy distillation.
\newblock \emph{Transactions on Machine Learning Research}, 2025.
\newblock ISSN 2835-8856.
\newblock URL \url{https://openreview.net/forum?id=WfVXe88oMh}.

\bibitem[Villani(2009)]{villani2009optimal}
C{\'e}dric Villani.
\newblock \emph{Optimal Transport: Old and New}, volume 338 of
  \emph{Grundlehren der mathematischen Wissenschaften}.
\newblock Springer, 2009.
\newblock \doi{10.1007/978-3-540-71050-9}.
\newblock URL \url{https://doi.org/10.1007/978-3-540-71050-9}.

\bibitem[Wang et~al.(2025)Wang, Li, Mandlekar, Xu, Fan, Narang, Fan, Zhu,
  Balaji, Zhou, Liu, and Zeng]{wang2024onedp}
Zhendong Wang, Max Li, Ajay Mandlekar, Zhenjia Xu, Jiaojiao Fan, Yashraj
  Narang, Linxi Fan, Yuke Zhu, Yogesh Balaji, Mingyuan Zhou, Ming-Yu Liu, and
  Yu~Zeng.
\newblock One-step diffusion policy: Fast visuomotor policies via diffusion
  distillation.
\newblock In \emph{Proceedings of the 42nd International Conference on Machine
  Learning}, volume 267 of \emph{Proceedings of Machine Learning Research},
  pp.\  63399--63416. PMLR, 2025.
\newblock URL \url{https://proceedings.mlr.press/v267/wang25ba.html}.

\bibitem[{Xiaomi Robotics}(2026{\natexlab{a}})]{checkpoint}
{Xiaomi Robotics}.
\newblock {Xiaomi-Robotics-1-RoboCasa365}: checkpoint and evaluation
  instructions, 2026{\natexlab{a}}.
\newblock URL
  \url{https://huggingface.co/XiaomiRobotics/Xiaomi-Robotics-1-RoboCasa365}.

\bibitem[{Xiaomi Robotics}(2026{\natexlab{b}})]{xiaomi0checkpoint}
{Xiaomi Robotics}.
\newblock {\mbox{Xiaomi-Robotics-0}-LIBERO}: checkpoint and model card,
  2026{\natexlab{b}}.
\newblock URL
  \url{https://huggingface.co/XiaomiRobotics/Xiaomi-Robotics-0-LIBERO}.

\bibitem[{Xiaomi Robotics Team} et~al.(2026){Xiaomi Robotics Team}, Guo, Jin,
  Li, Li, Li, Liu, Peng, Qin, Su, et~al.]{xiaomi}
{Xiaomi Robotics Team}, Jun Guo, Piaopiao Jin, Jason Li, Peiyan Li, Yingyan Li,
  Futeng Liu, Wanli Peng, Optimus Qin, Yifei Su, et~al.
\newblock {Xiaomi-Robotics-1: Scaling Vision-Language-Action Models with over
  100K Hours of Real-World Trajectories}.
\newblock \emph{arXiv preprint arXiv:2607.15330}, 2026.
\newblock URL \url{https://arxiv.org/abs/2607.15330}.

\end{thebibliography}

\clearpage
\appendix
\interlinepenalty=10000
\setlength{\textfloatsep}{20pt plus 2pt minus 4pt}
\setlength{\floatsep}{12pt plus 2pt minus 2pt}
\setlength{\intextsep}{12pt plus 2pt minus 2pt}
\section{Gaussian flow theory and feature analysis}
\label{app:gaussian-fm}

For a Gaussian target, the flow-matching velocity is affine in its input.
Fix the context $c$, write $\boldsymbol\mu=\boldsymbol\mu(c)$ and $\boldsymbol\Sigma=\boldsymbol\Sigma(c)$, and assume $\boldsymbol\Sigma$ is positive definite.
Let $\mathbf M\sim\mathcal N(\boldsymbol\mu,\boldsymbol\Sigma)$ and $\mathbf Z\sim\mathcal N(\mathbf 0,\mathbf I)$ be independent, with $\mathbf X_t=(1-t)\mathbf Z+t\mathbf M$.
The velocity field is
\begin{equation}
    \mathbf v(\mathbf x,t,c)
    =\mathbb E[\mathbf M-\mathbf Z\mid\mathbf X_t=\mathbf x,c].
    \label{eq:fm-conditional-velocity}
\end{equation}
The interpolated action and target velocity are jointly Gaussian, with
\begin{equation}
    \begin{aligned}
        \mathbb E[\mathbf X_t\mid c]         & =t\boldsymbol\mu,
                                             & \operatorname{Cov}(\mathbf X_t\mid c)                     & =\mathbf C_t=(1-t)^2\mathbf I+t^2\boldsymbol\Sigma, \\
        \mathbb E[\mathbf M-\mathbf Z\mid c] & =\boldsymbol\mu,
                                             & \operatorname{Cov}(\mathbf M-\mathbf Z,\mathbf X_t\mid c) & =t\boldsymbol\Sigma-(1-t)\mathbf I.
    \end{aligned}
\end{equation}
For jointly Gaussian vectors $\mathbf Y$ and $\mathbf X$, with $\operatorname{Cov}(\mathbf X)$ invertible, the conditional mean is
\begin{equation*}
    \mathbb E[\mathbf Y\mid\mathbf X=\mathbf x]
    =\mathbb E[\mathbf Y]
    +\operatorname{Cov}(\mathbf Y,\mathbf X)
    \operatorname{Cov}(\mathbf X)^{-1}
    \bigl(\mathbf x-\mathbb E[\mathbf X]\bigr).
\end{equation*}
Applying this formula at fixed $c$, with $\mathbf Y=\mathbf M-\mathbf Z$ and $\mathbf X=\mathbf X_t$, gives
\begin{align}
    \mathbf v(\mathbf x,t,c)
     & =\boldsymbol\mu+\mathbf A_t(\mathbf x-t\boldsymbol\mu),
    \label{eq:gar_affine_flow}                                         \\
    \mathbf A_t
     & =\bigl[t\boldsymbol\Sigma-(1-t)\mathbf I\bigr]\mathbf C_t^{-1}.
    \nonumber
\end{align}
For $0<t<1$, solving for the covariance gives
\begin{equation}
    \boldsymbol\Sigma
    =\frac{1-t}{t}
    \bigl[\mathbf I-t\mathbf A_t\bigr]^{-1}
    \bigl[\mathbf I+(1-t)\mathbf A_t\bigr].
    \label{eq:gar_moment_recovery}
\end{equation}
Thus the field is affine in $\mathbf x$, with coefficients determined by the target mean, covariance and interpolation time.

At the noise endpoint, $\mathbf X_0=\mathbf Z$ is independent of $\mathbf M$ given $c$, so
\begin{equation}
    \mathbf v(\mathbf x,0,c)=\mathbb E[\mathbf M\mid c]-\mathbf x,
    \qquad
    \mathbf x+\mathbf v(\mathbf x,0,c)=\mathbb E[\mathbf M\mid c].
    \label{eq:fm-initial-mean}
\end{equation}
This identity also holds for non-Gaussian targets with finite second moments.
A single Euler step of size one from $t=0$ therefore returns the conditional mean.

Covariance affects the Gaussian velocity at intermediate times but is absent at $t=0$.
To motivate predicting variance from time-zero features, Appendix~\ref{app:feature-time} compares the teacher's internal features across these times at fixed context and noisy input.

\subsection{Diagonal variance recovery}
\label{app:diagonal-recovery}

For diagonal $\Sigma(c)$, each velocity coordinate depends only on the corresponding action coordinate.
Given the mean $\mu(c)$, a velocity query at $0<t<1$ determines
\begin{equation}
    \alpha_i(t,c)
    =\frac{v_i(m_t,t\mid c)-\mu_i(c)}{m_{t,i}-t\mu_i(c)}
    =\frac{t\sigma_i^2(c)-(1-t)}{(1-t)^2+t^2\sigma_i^2(c)},
    \label{eq:gar_diagonal_query}
\end{equation}
provided $m_{t,i}\neq t\mu_i(c)$.
Multiplying by the denominator and collecting the terms containing $\sigma_i^2(c)$ gives
\begin{equation*}
    t\bigl[1-t\alpha_i(t,c)\bigr]\sigma_i^2(c)
    =(1-t)\bigl[1+(1-t)\alpha_i(t,c)\bigr].
\end{equation*}
Hence,
\begin{equation*}
    \sigma_i^2(c)
    =\frac{(1-t)\bigl[1+(1-t)\alpha_i(t,c)\bigr]}
    {t\bigl[1-t\alpha_i(t,c)\bigr]}.
\end{equation*}
Thus, once the mean is known, one intermediate-time velocity query recovers all diagonal variances, provided every query coordinate satisfies the condition above.
A nondegenerate Gaussian query satisfies these conditions with probability one.

For full covariance, the coordinates are coupled through $A_t(c)$.
One velocity query determines only the matrix--vector product
\begin{equation}
    v(m_t,t\mid c)-\mu(c)
    =A_t(c)\bigl(m_t-t\mu(c)\bigr).
    \label{eq:gar_full_covariance_query}
\end{equation}
Recovering the full matrix generally requires additional independent query directions or the velocity Jacobian, $\nabla_{m_t}v=A_t(c)$.

\subsection{Feature changes near the initial time}
\label{app:feature-time}

At $t=0$, the ideal velocity gives us the mean but says nothing about variance.
The internal features may still carry this information, since at nearby times they support a velocity that depends on variance.
If these features change little with time, we may be able to read out both moments from the same evaluation.
This motivates comparing the features at $t=0$, $0.01$, and $0.1$, keeping the context and noisy action input fixed.

For each teacher, we use 32 saved validation observations from distinct Human300 tasks and three Gaussian noise draws per observation, giving 96 pairs for each comparison with $t=0$.
GR00T uses the first validation observation from each of 32 task indices spaced evenly across the 300-task bank.
Xiaomi uses 32 evenly spaced entries from an existing 50-task validation bank, each containing the first validation observation for that task and its native four-frame history.
The task subsets differ between teachers; every within-teacher comparison holds images, instruction, robot state and the exact noisy action tensor fixed, changing only time.
The noise seeds are 10817--10819 for GR00T and 47017--47019 for Xiaomi, including each deployed readout's fixed query.

We capture the action-transformer output $h$ and the readout features $u$.
The deployed moment readouts use $u$ for both teachers.
All sixteen action-token feature vectors are flattened before computing
\begin{equation}
    \operatorname{cos}(f_0,f_t)
    =\frac{\langle f_0,f_t\rangle}{\|f_0\|_2\|f_t\|_2},
    \qquad
    \Delta_{\mathrm{rel}}
    =\frac{\|f_t-f_0\|_2}{\|f_0\|_2},
    \qquad
    r=\frac{\|f_t\|_2}{\|f_0\|_2}.
\end{equation}
To check whether a common feature offset dominates cosine similarity, we also subtract the across-observation mean separately for each time and noise seed before computing centered cosine similarity.
Table~\ref{tab:feature-time} reports descriptive means across the 96 pairs.

\begin{table}[htbp]
    \small
    \setlength{\tabcolsep}{5pt}
    \caption{Feature changes with time at an identical noisy input.
        Each row averages 32 validation observations and three noise draws, relative to $t=0$.
        Norm change is $100(r-1)$; centered cosine removes the mean feature at each time and noise seed.}
    \label{tab:feature-time}
    \begin{center}
        \begin{tabular}{llrrrrr}
            \toprule
            Teacher & Features & $t$  & Cosine  & $\Delta_{\mathrm{rel}}$ (\%) & Norm change (\%) & Centered cosine \\
            \midrule
            GR00T   & $h$      & 0.01 & 0.99993 & 1.68                         & +1.16            & 0.99958         \\
                    & $u$      & 0.01 & 0.99997 & 1.29                         & +1.06            & 0.99961         \\
                    & $h$      & 0.10 & 0.99937 & 11.39                        & +10.76           & 0.99684         \\
                    & $u$      & 0.10 & 0.99981 & 11.12                        & +10.94           & 0.99635         \\
            \midrule
            Xiaomi  & $h$      & 0.01 & 0.99998 & 1.03                         & +0.76            & 0.99979         \\
                    & $u$      & 0.01 & 0.99999 & 1.01                         & +0.89            & 0.99982         \\
                    & $h$      & 0.10 & 0.99901 & 6.66                         & +4.84            & 0.99764         \\
                    & $u$      & 0.10 & 0.99936 & 9.73                         & +8.98            & 0.99778         \\
            \bottomrule
        \end{tabular}
    \end{center}
\end{table}

For each fixed observation and noisy input, feature directions remain highly similar across the tested times for both teachers.
The readout features $u$ mainly increase in magnitude.

The predicted velocities also remain nearly parallel across these times.
Between $t=0$ and $t=0.1$, their mean cosine similarity is $0.99987$ for GR00T and $0.99993$ for Xiaomi, measured over the twelve action channels used for execution.
Their norms increase by about $11\%$ for both teachers.

High cosine similarity alone does not establish preservation of variance information, which could occupy a small component of the representation.
The held-out moment-prediction experiments test that information's accessibility directly.

\FloatBarrier
\section{Experimental setup}
\label{app:setting}

\subsection{RoboCasa365}

RoboCasa365 is a simulation benchmark with 365 household tasks across 2,500 kitchen environments~\citep{robocasa}.
A mobile robot with an arm performs basic skills, such as opening a drawer, and multistep activities, such as preparing coffee.
Both teachers use released RoboCasa365 checkpoints trained on Human300 demonstrations.
Xiaomi uses the \href{https://huggingface.co/XiaomiRobotics/Xiaomi-Robotics-1-RoboCasa365}{Xiaomi-Robotics-1-RoboCasa365 checkpoint} with five flow updates~\citep{checkpoint,xiaomi}.
GR00T N1.5 uses the \path{gr00t_n1-5/multitask_learning/checkpoint-120000} checkpoint from \href{https://huggingface.co/robocasa/robocasa365_checkpoints}{the RoboCasa365 checkpoint release}, with four flow updates~\citep{gr00t}.

Evaluation covers the 50-task target suite on the \texttt{pretrain} environment split: 18 Atomic-Seen, 16 Composite-Seen and 16 Composite-Unseen tasks.
Xiaomi inputs include four observations spaced two simulator steps apart, the released 0.95 image crop and the robot state; GR00T uses its native current-frame inputs and preprocessing.
Both policies execute 16 actions before replanning.
The evaluator retains native task horizons, scene randomization and persistent workers.
The main comparison uses 50 episodes per task and policy.
Context selection and readout fitting are specified in Appendices~\ref{app:context-sampling} and~\ref{app:moment-readouts}.

The benchmark comparisons use independently drawn physical scenes, while the separate fixed-scene diagnostics restore saved starts.
Episode indices alone do not reproduce the native scene draws because policies use different numbers of servers and persistent GPU random streams.

Raw one-step execution tests how well the teacher performs with a single flow update: it uses GAR's fixed noise input and the teacher's native action decoding, without fitted readouts or added action noise.
Each GAR setting in Table~\ref{tab:gar_success} uses one fitted readout.

\begin{figure}[!htbp]
    \centering
    \includegraphics[width=\linewidth]{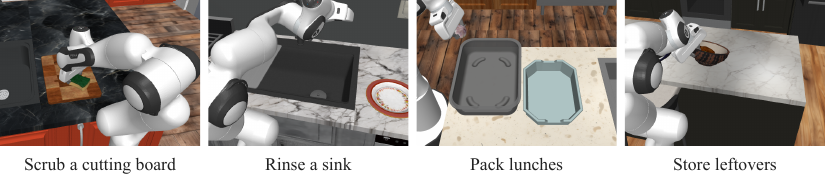}
    \caption{\textbf{One-pass Gaussian control on RoboCasa365.}
        Selected successful executions using GR00T full-covariance readouts (left two) and Xiaomi diagonal readouts (right two).
        Each action chunk uses one action-transformer evaluation.}
    \label{fig:robocasa_teaser}
\end{figure}

\subsection{LIBERO-Long}
\label{app:libero-setting}

We evaluate the released \mbox{Xiaomi-Robotics-0} LIBERO checkpoint~\citep{xiaomi0checkpoint} using five flow steps for the teacher and one action-transformer evaluation for GAR and raw one-step execution.
All policies execute ten actions before replanning, with ten initial settling steps and a 520-step control limit.
Evaluation uses fifty common official initial states per task, giving 500 episodes per policy.

\FloatBarrier
\section{Readout fitting}
\label{app:readout-fitting}

\subsection{RoboCasa365 context sampling}
\label{app:context-sampling}
We sample contexts to cover every Human300 task and different stages of its demonstrations.
For each of the 300 tasks, we randomly select five of the first 100 human demonstrations without replacement and take four evenly spaced interior states from each, giving 6,000 fitting contexts.
For a demonstration of $L$ frames, the zero-based indices are $\lfloor 6+j(L-23)/(K+1)\rfloor$, $j=1,\ldots,K$, with $K=4$ for fitting.
These indices spread the states between frames $6$ and $L-17$, leaving room for the observation history and action chunk.
Both teachers use the same contexts, each with its native observation and history inputs.

Tuning and evaluation use separate demonstrations from fitting and from each other.
Two additional demonstrations per task supply two states each ($K=2$); we assign one demonstration to tuning and the other to evaluation, giving 600 contexts in each set.
Demonstration selection uses seed 66017 plus the task index; the tuning/evaluation assignment uses seed 11717.
The evaluation bank has also appeared in earlier exploratory analyses.

Eligible training states are frames $6$ through $L_d-17$, inclusive, of demonstration $d$, where $L_d$ is its number of frames.
Among the first 100 demonstrations per task, excluding those reserved for tuning and evaluation leaves 29,400 demonstrations, giving
\[
    N=\sum_{d=1}^{29{,}400}(L_d-22)=26{,}612{,}851
\]
eligible training states.
The 6,000 fitting contexts therefore constitute $100\times6{,}000/N\approx0.02\%$ of this pool.

To test the effect of more fitting data, the 24,000-context fits add fifteen training demonstrations per task using the same four-state rule.
The original fitting, tuning and evaluation contexts are preserved, and command readouts use the same 6,000-context fit for both fitting-set sizes.

At each context, 384 teacher samples determine the empirical mean and regularized covariance target.
GR00T has 144 varying continuous coordinates; its remaining 16 constant coordinates are kept at zero.
Xiaomi uses all 160 normalized continuous coordinates.

\subsection{RoboCasa365 readout fitting}
\label{app:moment-readouts}

Numerically inverting the variance-recovery identity can be sensitive to errors near $t=0$ or $m_{t,i}=t\mu_i(c)$.
GAR therefore fits empirical teacher moments estimated from samples.

Let $d_m$ and $d_b$ denote the numbers of continuous and binary coordinates in an action chunk.
The mean and command readouts act on the frozen readout features at each action position.
The mean readout adds an affine correction to the continuous coordinates of the teacher's one-step prediction.
The covariance readout averages features across positions and applies a single 128-unit GELU hidden layer.
For full covariance, it predicts a Cholesky factor with positive diagonal entries; the diagonal variant predicts positive variances.
Full covariance includes dependence between continuous coordinates at different chunk positions.
Command latents are independent of one another and of continuous actions.

\paragraph{Gaussian latents and Bernoulli commands.}
\label{app:gaussian-bernoulli}
Thresholding a unit-variance Gaussian can represent any Bernoulli probability $p\in(0,1)$.
For each binary command, we sample $z_j=\eta_{\theta,j}(c)+\xi_j$ with $\xi_j\sim\mathcal{N}(0,1)$ and set $b_j=\mathbf{1}\{z_j\geq0\}$.
Its activation probability is
\begin{equation}
    \begin{aligned}
        p_{\theta,j}(c)
         & =\Pr(b_j=1\mid c)
        =\Pr\!\left(\xi_j\geq-\eta_{\theta,j}(c)\right) \\
         & =1-\Phi\!\left(-\eta_{\theta,j}(c)\right)
        =\Phi\!\left(\eta_{\theta,j}(c)\right),
    \end{aligned}
    \label{eq:gar_command_probabilities}
\end{equation}
where $\Phi$ is the standard normal cumulative distribution function.
Thus $b_j\mid c\sim\operatorname{Bernoulli}(p_{\theta,j}(c))$.
Choosing $\eta=\Phi^{-1}(p)$ gives the desired probability, with zero and one obtained as limits.
Fixing the latent variance to one therefore does not restrict individual command probabilities.
Binary cross-entropy against teacher command frequencies trains the probabilities of the executed commands.

The marginal result also holds for correlated Gaussian latents.
For $z\mid c\sim\mathcal{N}(\eta_\theta(c),C(c))$, with $C_{jj}(c)>0$, thresholding gives
\begin{equation}
    \begin{aligned}
        b_j & =\mathbf{1}\{z_j\geq0\}, \\
        p_{\theta,j}(c)
         & =1-\Phi\!\left(-\frac{\eta_{\theta,j}(c)}{\sqrt{C_{jj}(c)}}\right)
        =\Phi\!\left(\frac{\eta_{\theta,j}(c)}{\sqrt{C_{jj}(c)}}\right).
    \end{aligned}
    \label{eq:gar_correlated_command_probabilities}
\end{equation}
Each binary marginal is therefore fully specified by its mean, regardless of correlations between latents.
The joint binary probabilities depend on those correlations and generally require Gaussian orthant integrals.
GAR uses independent unit-variance command latents, also independent of continuous actions, yielding the factorized likelihood in Equation~\eqref{eq:gar_likelihood}.

\paragraph{Offline teacher statistics.}
We select a small set of contexts $\mathcal{C}$ and draw $N$ teacher action chunks $a^{(i)}=(m^{(i)},b^{(i)})$ at each context.
Binary commands are obtained using the teacher's execution thresholds.
We estimate continuous-action moments and command frequencies as
\begin{align}
    \widehat\mu_c
     & = \frac{1}{N}\sum_{i=1}^{N}m^{(i)},
    \label{eq:gar_teacher_statistics}        \\
    \widehat S_c
     & = \frac{1}{N}\sum_{i=1}^{N}
    (m^{(i)}-\widehat\mu_c)
    (m^{(i)}-\widehat\mu_c)^\top,
    \nonumber                                \\
    \widehat p_{c,j}
     & = \frac{1}{N}\sum_{i=1}^{N}b_j^{(i)}.
    \nonumber
\end{align}
We shrink $\widehat S_c$ by 5\% toward its diagonal and floor eigenvalues at the larger of $10^{-5}$ times its mean diagonal and $10^{-10}$ to obtain $\widehat\Sigma_c$.
This regularization weakens correlations and prevents nearly singular covariance targets; it is separate from the readout loss.
Shrinkage toward the diagonal preserves marginal variances, and eigenvalue flooring can only increase them.
These statistics are cached together with the readout features and one-step teacher predictions.

\paragraph{Moment matching.}
We fit the affine mean readout by least squares:
\begin{equation}
    \mathcal{L}_{\mu}(\theta)
    =
    \mathbb{E}_{c\sim\mathcal{C}}\!\left[
        \frac{1}{d_m}
        \left\|
        \mu_\theta(c)-\widehat\mu_c
        \right\|_2^2
        \right].
    \label{eq:gar_mean_loss}
\end{equation}
We then fit the covariance readout while keeping the mean and command readouts fixed.
Covariance targets are centered on the empirical teacher mean, so covariance fitting is independent of mean-prediction error.

To account for differences in scale between action coordinates, let $D=\operatorname{diag}(s_1,\ldots,s_{d_m})$, where $s_k^2$ is the average target variance of coordinate $k$ over the original 6,000 fitting contexts, for both fitting-set sizes.
For full covariance, we minimize
\begin{equation}
    \begin{aligned}
        \mathcal{L}_{\Sigma}(\theta)
         & =
        \mathbb{E}_{c\sim\mathcal{C}}\!\Bigg[
        \frac{1}{d_m^2}
        \left\|
        D^{-1}
        \bigl(\Sigma_\theta(c)-\widehat\Sigma_c\bigr)
        D^{-1}
        \right\|_F^2
        \\
         & \hspace{3em}
        + \lambda_{\mathrm{var}}\,
        \mathcal{R}_{\mathrm{var}}(c)
        \Bigg],
    \end{aligned}
    \label{eq:gar_distillation_loss}
\end{equation}
where
\begin{equation}
    \mathcal{R}_{\mathrm{var}}(c)
    =
    \frac{1}{d_m}
    \sum_{k=1}^{d_m}
    \left(
    \log
    \frac{\Sigma_{\theta,kk}(c)}
    {\widehat\Sigma_{c,kk}}
    \right)^2,
    \label{eq:gar_variance_regularization}
\end{equation}
and $\lambda_{\mathrm{var}}=0.1$.
The first term matches covariance entries after coordinate-wise normalization, balancing action coordinates with different scales.
The logarithmic term also gives weight to errors at contexts with small target variances: doubling and halving a variance incur the same penalty, regardless of its scale.
For the diagonal variant, the first term averages normalized squared errors over the $d_m$ diagonal entries only.
This training penalty is separate from the shrinkage and eigenvalue flooring used to estimate covariance targets.

\paragraph{Covariance optimization.}
Both covariance variants use AdamW~\citep{loshchilov2019decoupled} with learning rate $10^{-3}$, weight decay $10^{-2}$, batch size 64, gradient clipping at norm one and seed 17.
We evaluate the corresponding moment loss on the 600 tuning contexts every 1,000 updates and select its lowest-loss checkpoint.
From update 12,000, four tuning evaluations without an improvement exceeding the larger of $10^{-4}$ in relative loss and $10^{-7}$ in absolute loss reduce the learning rate by a factor of 0.3, down to $10^{-6}$.
At this minimum rate, eight evaluations without such an improvement stop fitting.
Appendix~\ref{app:moment-diagnostics} reports the selected updates and convergence curves.

\paragraph{Command fitting.}
The command readout learns the teacher's activation probabilities by minimizing binary cross-entropy against the sampled command frequencies:
\begin{equation}
    \mathcal{L}_{b}(\theta)
    =
    \mathbb{E}_{c\sim\mathcal{C}}\!\left[
        \frac{1}{d_b}
        \sum_{j=1}^{d_b}
        \operatorname{BCE}\!\left(
        \widehat p_{c,j},
        \Phi(\eta_{\theta,j}(c))
        \right)
        \right],
    \label{eq:gar_command_loss}
\end{equation}
where $\operatorname{BCE}$ denotes binary cross-entropy.
This objective fits command activation probabilities through the Gaussian latent means.
For each teacher, the 6,000- and 24,000-context GAR variants share the same command readout, trained on 6,000 contexts.
During command training, the Bernoulli term above is combined with a Gaussian KL term for continuous motion.
We use 3,000 AdamW~\citep{loshchilov2019decoupled} updates, batch size 64, learning rate $10^{-4}$, weight decay $10^{-5}$ and gradient clipping at norm one, with checkpoint selection by validation motion-and-command likelihood.
The mean and covariance readouts described above replace the continuous-action predictors from this joint fit; the command parameters stay fixed.
All readout fitting uses cached features and teacher statistics, without further teacher evaluations or action sampling.

\subsection{LIBERO-Long readout fitting}
\label{app:libero-fitting}

For each task, thirty demonstrations supply twenty contexts each, giving 6,000 fitting contexts overall.
Separate sets of five demonstrations per task supply twelve contexts each for tuning and moment evaluation, giving 600 contexts in each set.
At each context, 384 teacher samples estimate continuous-action moments and gripper-command frequencies.

GAR adds an affine correction to the teacher's one-step continuous-action prediction and predicts diagonal covariance through a pooled 128-unit GELU readout.
The teacher networks remain frozen.

We use a probit readout with shared feature weights and a separate bias for each of the ten action positions.
Its coefficients were fitted by binary cross-entropy on the training contexts and selected by tuning cross-entropy.

\subsection{Fitting and preparation cost}
\label{app:fitting-cost}

For Xiaomi, the main cost is preparing the teacher data.
Once the readout features and targets are cached, the recorded fitting times total about two minutes for either the diagonal or full-covariance readouts on 6,000 contexts.

\paragraph{Fitting cost.}
The Xiaomi 6,000-context affine mean is solved by CPU least squares in 3.7 seconds.
Its diagonal and full covariance fits take 94.4 and 100.3 seconds on an RTX 4090.
The initial joint motion-and-command fit that supplies the fixed command readout adds 16.7 seconds on an RTX 4090, including optimization, tuning evaluations and checkpoint saving but excluding untimed data loading and setup.
The recorded fitting times therefore total 114.8 seconds for diagonal GAR and 120.7 seconds for full covariance; teacher sampling and feature extraction are counted separately below.

\paragraph{Preparation cost.}
The Xiaomi preparation measurements cover a cache of 7,200 fitting, tuning and evaluation contexts, each with 512 teacher chunks (3,686,400 chunks in total).
At each context, 384 chunks estimate the targets and the remaining 128 are reserved for density and variance comparisons.
\begin{table}[!htbp]
    \centering
    \small
    \caption{Preparation cost for Xiaomi's cache of 7,200 contexts and 3,686,400 teacher chunks.
        The measurements exclude teacher pretraining and policy evaluation; checkpoint download was not timed.}
    \label{tab:gar-preparation-cost}
    \begin{tabular*}{\linewidth}{@{\extracolsep{\fill}}llr@{}}
        \toprule
        Preparation step & Resources & Time \\
        \midrule
        Download, extraction and data checks & CPU/network & 2.71 h wall time \\
        \addlinespace[2pt]
        Feature extraction and teacher sampling & 4 RTX 4090 GPUs & \shortstack[r]{3.84 GPU-hours total\\1.14 h longest parallel job} \\
        \addlinespace[2pt]
        Motion and command target estimation & 12 CPU workers & 6.9 s wall time \\
        \bottomrule
    \end{tabular*}
\end{table}

\FloatBarrier
\section{Gaussian approximation in physical action units}
\label{app:physical_action_scale}

\subsection{Shared procedure}

\paragraph{Samples and units.}
Both teachers use the same 300 saved observation identities, selecting one validation observation uniformly within each Human300 task bank.
Each observation supplies 1,024 teacher chunks, generated with FP32 action-network computation and Euler accumulation while retaining frozen vision-language conditioning.
Xiaomi uses its native four-frame conditioning and five flow steps; GR00T uses current-frame conditioning and four flow steps.
Binary commands remain unrestricted during sampling but are excluded from the continuous-action comparison.
The first 512 samples fit the maximum-likelihood Gaussian mean and covariance; the remaining 512 evaluate the approximation.
Both teachers decode translation and rotation outputs directly into controller inputs.
The controller clips each coordinate to $[-1,1]$, then scales translation by 50\,mm and rotation-vector coordinates by 0.5 radians, which we convert to degrees.
We fit before clipping and apply the controller mapping to both teacher and Gaussian commands before projecting them.

\paragraph{Sliced comparison.}
Sliced 1-Wasserstein distance compares the teacher and Gaussian distributions along many one-dimensional projections~\citep{villani2009optimal,bonneel2015sliced}.
Each projection combines the action coordinates into one number, allowing the comparison to detect dependence that separate coordinate comparisons can miss.
Along each direction, we measure how far command values must move to match the two distributions, then average these distances across directions.
Small shifts in command values therefore give small errors, whereas KL divergence can be large for small shifts between narrowly concentrated distributions.

Let $\theta_1,\ldots,\theta_L$ be the unit projection directions, and let $\widehat F^{-1}_{c,\ell}$ and $\widehat G^{-1}_{c,\ell}$ denote the empirical quantile functions of the projected teacher and Gaussian reference points at context $c$.
For the $n$ held-out teacher commands $x_i(c)$ in physical units, we define
\begin{equation}
    \begin{aligned}
        D(c) & = \frac{1}{L}\sum_{\ell=1}^{L}\int_0^1
        \left|\widehat F^{-1}_{c,\ell}(u)-\widehat G^{-1}_{c,\ell}(u)\right|\,du, \\
        A(c) & = \frac{1}{Ln}\sum_{\ell=1}^{L}\sum_{i=1}^{n}
        \left|\theta_\ell^\top x_i(c)\right|,
        \qquad R(c)=\frac{D(c)}{A(c)}.
    \end{aligned}
    \label{eq:physical_action_scale}
\end{equation}
The projected command magnitude $A(c)$ averages the absolute projections of the teacher commands, without centering them.
Thus $R(c)$ measures approximation error relative to command size, including any nonzero mean.
We compute this ratio within each context, then report its median across contexts; all observed denominators are nonzero.

For each action-group dimension, we use $n=512$ held-out teacher commands and $L=256$ unit directions sampled uniformly on the sphere, shared across teachers and covariance choices.
To reduce numerical error in each Gaussian reference, we use 2,048 points from a scrambled Sobol sequence~\citep{sobol1967distribution,matousek1998discrepancy}.
Sobol sequences are designed to cover the unit cube evenly; scrambling randomizes the points while preserving this balanced coverage.
We transform these points through the standard-normal quantile function, multiply by the fitted covariance's symmetric square root, and add the fitted mean.
The reference points undergo the same clipping, scaling and projection as the teacher commands.
Sorting the projected points lets us evaluate each quantile integral exactly for these empirical distributions.

\paragraph{Gaussian sampling controls.}
Finite samples produce approximation error even when the underlying distribution is Gaussian, so we use Gaussian sampling controls to measure this baseline.
At each context and for each group, we draw eight independent sets of 1,024 samples from the fitted Gaussian and repeat the 512/512 split, refitting and projected comparison.
Sampling and refitting use the same covariance family as the corresponding teacher comparison.
We average the eight control errors before dividing by the original teacher's projected command magnitude.
These controls measure finite-sample discrepancy and reference integration error; they are not subtracted as an unbiased correction.

Rotation uses Euclidean rotation-vector coordinates, not geodesic orientation distance.
These measurements describe command distributions, not positioning error or a count of density modes; their consequences for task performance are evaluated separately in Section~\ref{sec:experiments_performance}.

\subsection{Action-chunk comparison}
\label{app:chunk-physical-action-scale}

The full-chunk comparison asks how well a Gaussian captures the joint distribution of commands across all sixteen action positions (Table~\ref{tab:gar_gaussian_error}).
We group the translation commands into one 48-dimensional vector and the rotation commands into another.
Full covariance retains dependence within and across action positions; diagonal covariance retains the same fitted means and marginal variances but sets all off-diagonal entries to zero.
Both approximations fit teacher samples directly, independently of the learned GAR readouts.

We apply the shared procedure to the 48-dimensional translation and rotation groups.
Both covariance choices use the same projected teacher-command magnitude as denominator.

\begin{table}[htbp!]
    \centering
    \small
    \caption{Gaussian approximation of translation and rotation across sixteen-action chunks at 300 fixed observations per teacher.
        Each group has 48 coordinates.
        Entries are median relative sliced 1-Wasserstein errors (\%), normalized by the teacher's projected command magnitude; lower is better.
        Controls repeat fitting and evaluation on samples from the corresponding full or diagonal Gaussian (Appendix~\ref{app:physical_action_scale}).
    }
    \label{tab:gar_gaussian_error}
    \begin{center}
        \begin{tabular}{@{}llrrrr@{}}
            \toprule
                    &              & \multicolumn{2}{c}{Full covariance} & \multicolumn{2}{c}{Diagonal covariance}                             \\
            \cmidrule(lr){3-4}\cmidrule(l){5-6}
            Teacher & Action group & Error (\%)                          & Control (\%)                            & Error (\%) & Control (\%) \\
            \midrule
            Xiaomi  & Translation  & 0.8                                 & 0.7                                     & 1.8        & 0.7          \\
                    & Rotation     & 1.8                                 & 1.6                                     & 3.0        & 1.5          \\
            \addlinespace
            GR00T   & Translation  & 4.6                                 & 3.3                                     & 9.1        & 3.4          \\
                    & Rotation     & 9.5                                 & 6.3                                     & 19.9       & 6.4          \\
            \bottomrule
        \end{tabular}
    \end{center}
\end{table}

\begin{figure}[htbp]
    \centering
    \includegraphics[width=\linewidth]{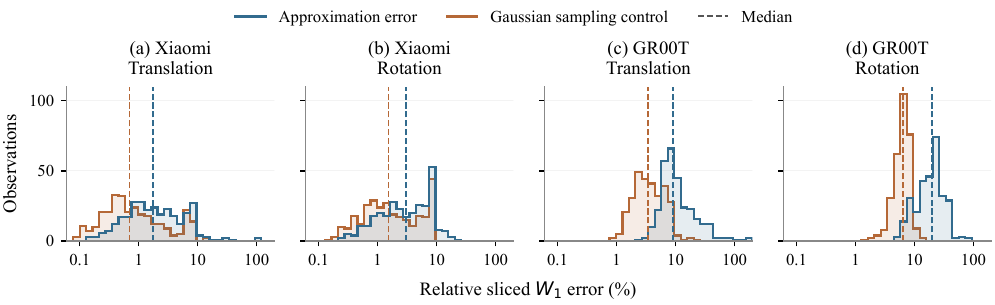}
    \caption{Diagonal-covariance Gaussian approximation under the same procedure as Figure~\ref{fig:gar_gaussian_error}, with 300 observations per panel.
        Approximation errors and Gaussian sampling controls use the same logarithmic bins and axis limits as the full-covariance comparison.
        Dashed lines mark medians, and all observations are retained.}
    \label{fig:gar_gaussian_error_diagonal}
\end{figure}

\begin{table}[htbp]
    \centering
    \small
    \setlength{\tabcolsep}{5pt}
    \caption{Physical sliced errors and upper-tail relative errors for the sixteen-action comparison on 300 contexts per teacher.
        Physical errors are medians across contexts, in millimetres for translation and degrees for rotation-vector commands.
        The last two columns give 90th percentiles of relative error (\%); lower values are better.
        These are descriptive summaries without confidence intervals; Table~\ref{tab:gar_gaussian_error} gives median relative errors and Gaussian controls.}
    \label{tab:chunk-physical-action-scale}
    \begin{tabular}{@{}llrrrr@{}}
        \toprule
                &            & \multicolumn{2}{c}{Physical error} & \multicolumn{2}{c}{Relative error, 90th percentile}                          \\
        \cmidrule(lr){3-4}\cmidrule(l){5-6}
        Teacher & Covariance & Translation                        & Rotation                                            & Translation & Rotation \\
        \midrule
        Xiaomi  & Full       & 0.0590                             & 0.0071                                              & 7.6         & 8.8      \\
                & Diagonal   & 0.1207                             & 0.0119                                              & 7.7         & 8.8      \\
        \addlinespace
        GR00T   & Full       & 0.4054                             & 0.0726                                              & 14.9        & 19.3     \\
                & Diagonal   & 0.9867                             & 0.1316                                              & 29.2        & 35.4     \\
        \bottomrule
    \end{tabular}
\end{table}

\paragraph{Numerical checks and scope.}
We vary the number of projection directions and Gaussian reference points to check numerical accuracy in the 48-dimensional groups.
Using 64, 128 or 256 directions changes aggregate median relative errors by at most 0.15 percentage points with full covariance and 0.75 with diagonal covariance.
On twelve matched contexts per teacher, increasing the reference to 8,192 points with an independent scramble changes individual relative errors by at most 0.59 and 0.24 percentage points, respectively.

The chunk comparison includes temporal dependence within translation and within rotation, while translation--rotation coupling and other commands remain outside its scope.
The physical distances concern projected command distributions, not accumulated displacement or orientation errors.

\FloatBarrier
\paragraph{Effect of covariance.}
Removing correlations increases median approximation errors for both teachers (Table~\ref{tab:gar_gaussian_error}).
The 90th-percentile relative errors remain nearly unchanged for Xiaomi and roughly double for GR00T (Table~\ref{tab:chunk-physical-action-scale}), showing a larger effect of removing correlations in GR00T's upper-tail errors.

\section{Moment recovery}
\label{app:moment-recovery}

Small moment errors relative to action magnitude coexist with substantial relative errors in individual variances (Section~\ref{sec:experiments_moments}).
We examine these errors more closely, compare them with uncertainty in the teacher estimates, and test how they change with the fitting loss.

\subsection{Moment prediction and convergence}
\label{app:moment-diagnostics}

\begin{figure}[!htbp]
    \centering
    \includegraphics[width=\linewidth]{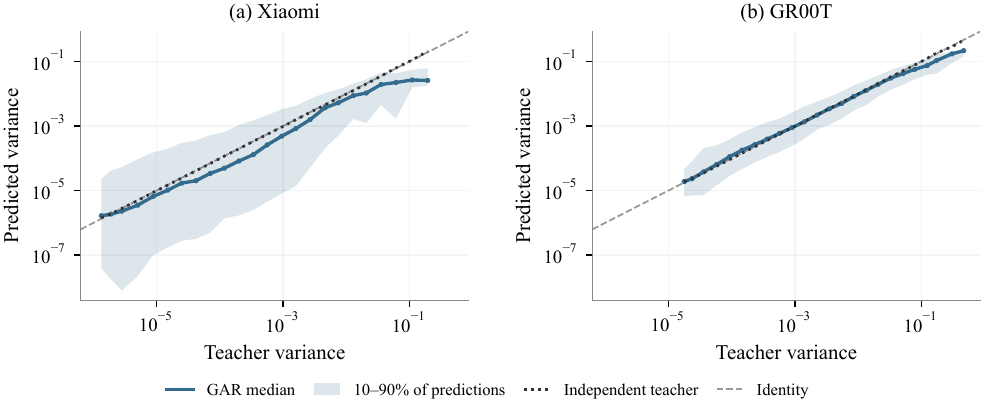}
    \caption{Variance predictions from diagonal GAR fitted on 6,000 contexts, evaluated on 600 held-out contexts per teacher.
        Predictions track teacher variances but underestimate the largest values, especially for Xiaomi.
        Curves show medians within logarithmic teacher-variance bins; shading spans the 10th--90th percentiles of predictions, not confidence intervals.
        The dotted curve compares variance estimates from 128 independent teacher samples against 384-sample targets, providing a finite-sampling reference.
        The following diagnostics give all predictions and upper-tail errors.}
    \label{fig:gar-variance-calibration}
\end{figure}

\paragraph{Errors relative to action magnitude.}
The SD metric in Table~\ref{tab:gar_moment_fit} measures marginal-scale errors relative to the magnitude of the teacher's mean action chunk.
For diagonal Gaussian sampling, holding the mean fixed and sharing a standard Gaussian noise vector $\epsilon$ gives
\begin{equation}
    \sqrt{\mathbb{E}_\epsilon\left[
            \|\bigl(\sigma_\theta(c)-\widehat\sigma_c\bigr)\odot\epsilon\|_2^2
            \right]}
    = \|\sigma_\theta(c)-\widehat\sigma_c\|_2.
\end{equation}
The numerator is therefore the root mean square change in the sampled action due to SD prediction error, measured in the teacher's fitted action coordinates.
For full-covariance readouts, this marginal metric omits correlation errors.

Table~\ref{tab:gar_moment_tail} complements the medians and interquartile ranges in Table~\ref{tab:gar_moment_fit} with 90th-percentile errors, showing how the errors grow toward the upper end of the distribution.
For the 6,000-context diagonal readouts, SD errors at this percentile reach 16.19\% for Xiaomi and 29.07\% for GR00T.
All 600 evaluation means per teacher have nonzero norm; no contexts are excluded and no denominators are floored.
Percentiles use linear interpolation between ordered values.

\begin{table}[htbp]
    \small
    \setlength{\tabcolsep}{5pt}
    \caption{90th-percentile moment errors on the same 600 evaluation contexts per teacher as Table~\ref{tab:gar_moment_fit}.
        All errors are percentages of the teacher mean's norm; lower is better.}
    \label{tab:gar_moment_tail}
    \begin{center}
        \begin{tabular}{ll*{3}{>{\centering\arraybackslash}p{8.5em}}}
            \toprule
            Teacher & Fitting contexts                  & Mean error (\%)
                    & \multicolumn{2}{c}{SD error (\%)}                                      \\
            \cmidrule(lr){4-5}
                    &                                   &                 & Diagonal & Full  \\
            \midrule
            Xiaomi  & 6,000                             & 13.63           & 16.19    & 10.65 \\
            Xiaomi  & 24,000                            & 13.40           & 13.90    & 10.34 \\
            GR00T   & 6,000                             & 20.08           & 29.07    & 25.26 \\
            GR00T   & 24,000                            & 19.44           & 25.86    & 24.65 \\
            \bottomrule
        \end{tabular}
    \end{center}
\end{table}

\paragraph{Diagonal variance accuracy.}
The readouts underestimate the largest variances beyond the discrepancy between teacher sample estimates.
Among the largest 1\% of target variances, median prediction-to-target ratios are 0.577 for Xiaomi and 0.633 for GR00T with the 6,000-context diagonal readouts, compared with 0.976 and 1.001 for independent teacher estimates (Table~\ref{tab:gar_variance_calibration}).

At each evaluation context, the targets use regularized covariance estimates from 384 teacher action chunks, while the independent estimates use unbiased sample variances from 128 disjoint teacher chunks.
For each teacher, we pool all context--coordinate pairs and select the largest 1\% by target variance: 960 pairs for Xiaomi and 864 for GR00T.
We divide each independent variance estimate by its corresponding target and take the median of these individual ratios over the selected pairs.
GAR's ratios use the same targets and selected pairs, with the predicted variance in the numerator.

\begin{figure}[p]
    \centering
    \includegraphics[width=\linewidth]{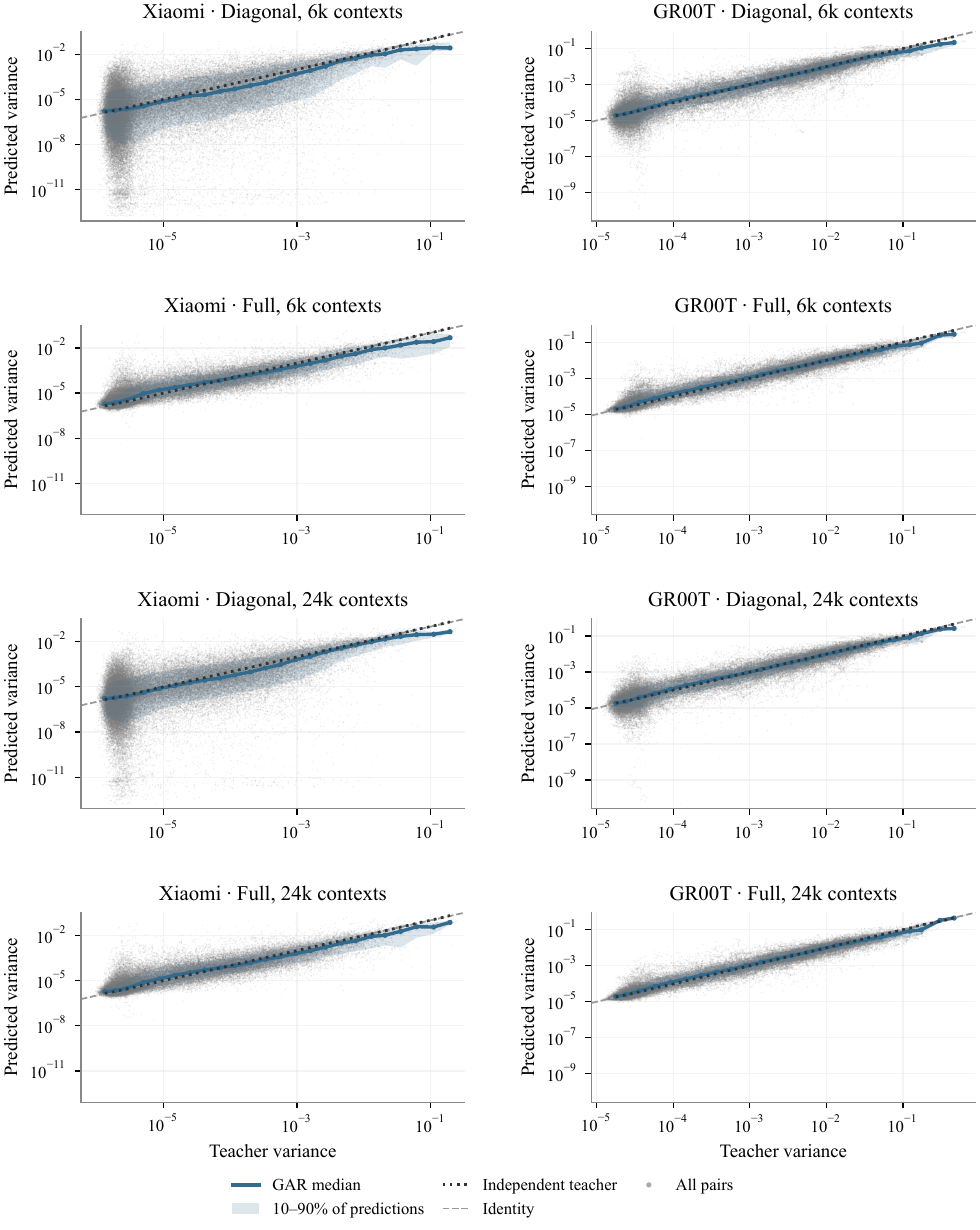}
    \caption{All held-out marginal-variance predictions from the deployed diagonal and full readouts.
        Gray points include every context--coordinate pair; curves and bands follow Figure~\ref{fig:gar-variance-calibration}.
        Each teacher uses common axis limits across its four panels, spanning the complete target and prediction ranges without clipping.
        GR00T's 16 structurally constant coordinates are excluded.
        Xiaomi uses normalized action coordinates and GR00T uses decoded coordinates.}
    \label{fig:gar-variance-all}
\end{figure}

Figure~\ref{fig:gar-variance-calibration} shows how predictions vary across the range of teacher variances.
It groups all 96,000 Xiaomi and 86,400 GR00T pairs into 24 equal-width bins in log variance, placing each point at the median target variance in its bin.
The independent teacher curve uses sample variances from 128 additional actions at each context.
To show the spread behind these summaries, Figure~\ref{fig:gar-variance-all} includes every prediction from all eight deployed checkpoints, including errors outside the central bands.

\begin{table}[htbp]
    \small
    \setlength{\tabcolsep}{5pt}
    \caption{Marginal variance accuracy on 600 evaluation contexts per teacher, relative to regularized 384-sample targets $\widehat v$.
        Independent teacher rows use unbiased variances from 128 disjoint teacher chunks per context; other rows use GAR predictions.
        Tail columns use the largest 1\% of target variances pooled across contexts and coordinates, reporting the median individual ratio $v/\widehat v$ (one is exact) and median relative error $|v/\widehat v-1|$ (lower is better).
        These targets exceed 0.00784 for Xiaomi and 0.0786 for GR00T, in their respective fitted action coordinates.
        Here $v$ denotes the independent estimate or GAR's predicted marginal variance.}
    \label{tab:gar_variance_calibration}
    \begin{center}
        \begin{tabular}{lllrr}
            \toprule
            Teacher & Fitting contexts & Estimate   & Tail ratio & Tail error (\%) \\
            \midrule
            Xiaomi & --- & \shortstack[l]{Independent\\teacher} & 0.976 & 10.9 \\
            \addlinespace[2pt]
            Xiaomi  & 6,000            & Diagonal   & 0.577      & 53.0            \\
            Xiaomi  & 6,000            & Full       & 0.503      & 56.5            \\
            Xiaomi  & 24,000           & Diagonal   & 0.669      & 45.5            \\
            Xiaomi  & 24,000           & Full       & 0.545      & 52.4            \\
            \midrule
            GR00T & --- & \shortstack[l]{Independent\\teacher} & 1.001 & 7.4 \\
            \addlinespace[2pt]
            GR00T   & 6,000            & Diagonal   & 0.633      & 37.5            \\
            GR00T   & 6,000            & Full       & 0.647      & 37.1            \\
            GR00T   & 24,000           & Diagonal   & 0.710      & 31.1            \\
            GR00T   & 24,000           & Full       & 0.667      & 36.5            \\
            \bottomrule
        \end{tabular}
    \end{center}
\end{table}

\begin{figure}[htbp]
    \centering
    \includegraphics[width=\linewidth]{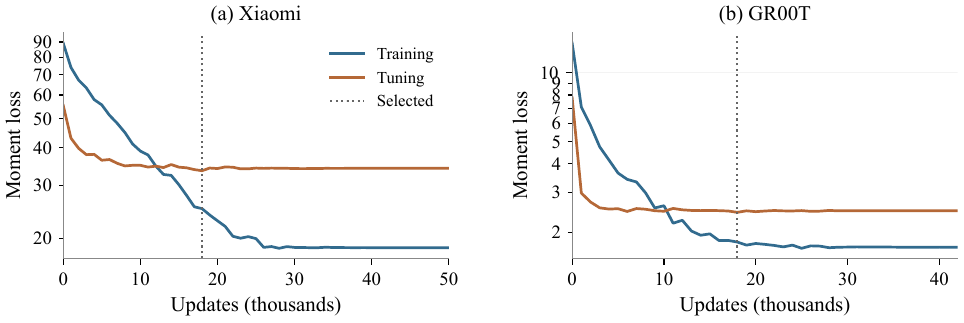}
    \caption{Convergence of the 6,000-context diagonal covariance readouts.
        Lower moment loss is better; vertical dotted lines mark the deployed checkpoints selected by tuning loss.
        Training curves use a fixed 1,200-context subset and tuning uses 600 contexts from separate demonstrations.
        Both fits select update 18,000 and stop at 50,000 updates for Xiaomi and 42,000 for GR00T.
        Each panel shows one fit, with the full recorded update range and a logarithmic loss axis.}
    \label{fig:gar-moment-convergence}
\end{figure}

\paragraph{Convergence.}
We track tuning loss through learning-rate reductions to check whether the fits have reached a plateau.
All eight covariance fits stop at a plateau, with selected checkpoints between 18,000 and 53,000 updates.%
\footnote{The 24,000-context Xiaomi covariance fits use CPUs, while the 6,000-context fits use a GPU, so their minibatch streams are not paired despite the common seed.}
Figure~\ref{fig:gar-moment-convergence} shows the loss curves for the two 6,000-context diagonal readouts, including continued fitting after the selected checkpoints.

\subsection{Teacher-sample uncertainty}
\label{app:sample-budget}

The teacher moments used as references are themselves estimated from finite samples.
We examine how the estimated mean changes with the number of teacher samples.

\paragraph{Mean estimates at different sample counts.}
For each of the original 6,000 Xiaomi fitting contexts, we estimate the continuous-action mean from nested subsets of 32, 128 and 384 cached teacher chunks.
The mean of $\|\widehat\mu_{c,N}-\widehat\mu_{c,384}\|_2^2/160$ across contexts is $1.23\times10^{-5}$ for $N=32$ and $2.24\times10^{-6}$ for $N=128$, in squared normalized-action units.
The 128 reserved teacher samples at each context are excluded from this comparison.

The mean squared difference from the 384-draw reference falls by a factor of 5.5 when increasing the sample count from 32 to 128.
The shared reference is itself estimated, so these differences measure stability under changes in sample count rather than error against the true teacher mean.

\subsection{Loss ablation for variance prediction}
\label{app:robust-variance}

\begin{figure}[htbp]
    \centering
    \includegraphics[width=\linewidth]{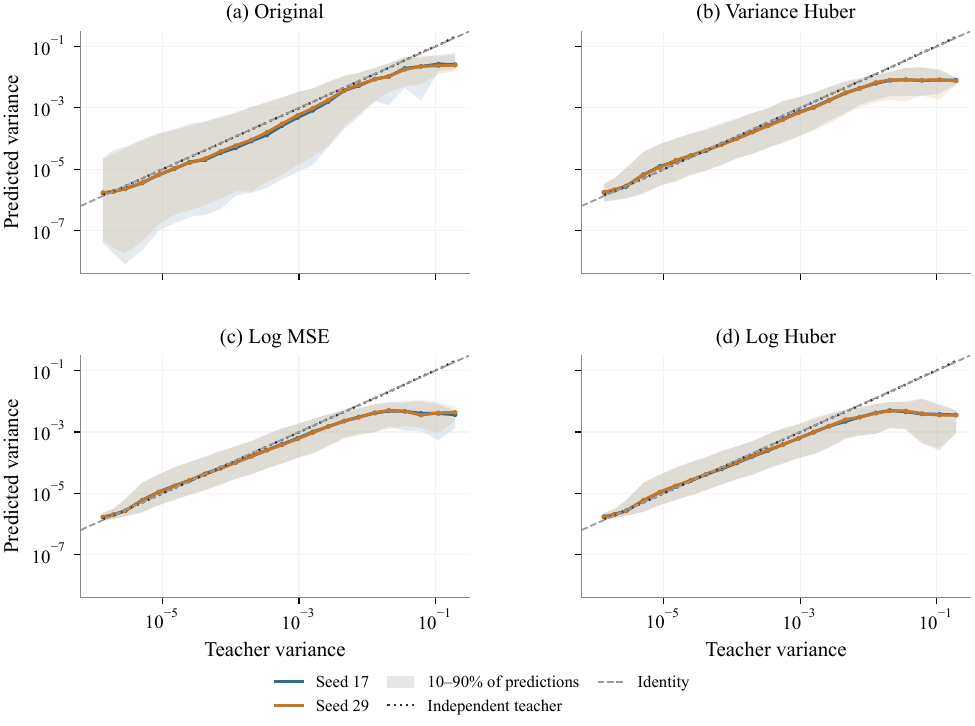}
    \caption{Effect of the fitting loss on Xiaomi variance prediction, with checkpoints selected by their respective tuning objectives.
        Both seeds use all 600 evaluation contexts and 160 coordinates per context.
        Curves show median predictions in the same 24 logarithmic bins; shading shows the 10th--90th percentiles for each seed, not confidence intervals.
        Axes are shared across panels and span the full target-variance range.
        The dotted curve uses 128 independent teacher samples, while targets use 384 samples.
        Log-space fitting improves typical predictions but increases underestimation of the largest variances.}
    \label{fig:robust-variance}
\end{figure}

We next test how much variance prediction depends on the fitting loss.
Changing the fitting loss reduces typical Xiaomi SD errors without changing the readout architecture.
The two log-space losses give similar improvements, while applying a Huber penalty to the original variance residual gives a smaller gain (Table~\ref{tab:robust-variance}).
These improvements come with stronger underestimation of the largest target variances (Figure~\ref{fig:robust-variance}).

\paragraph{Comparing fitting losses.}
The original loss combines squared variance residuals with a log-ratio penalty.
We compare two ways of changing this emphasis: reducing the influence of large variance residuals with a Huber penalty, and measuring multiplicative discrepancies in log space.
Let $r_{ck}=(v_{\theta,k}(c)-\widehat v_{ck})/s_k^2$ be the variance residual normalized by the training mean variance, and let $\ell_{ck}=\log(v_{\theta,k}(c)/\widehat v_{ck})$ be its log ratio.
The four losses average the following penalties over contexts and coordinates:
\begin{equation}
    \begin{aligned}
        \text{Original:}\quad & r_{ck}^{2}+0.1\ell_{ck}^{2},
                              & \qquad \text{Variance Huber:}\quad & h(r_{ck})+0.1\ell_{ck}^{2}, \\
        \text{Log MSE:}\quad  & \ell_{ck}^{2},
                              & \qquad \text{Log Huber:}\quad      & h(\ell_{ck}),
    \end{aligned}
\end{equation}
where $h(u)=u^2$ for $|u|\leq 1$ and $h(u)=2|u|-1$ otherwise.

\paragraph{Matched fits.}
To isolate the effect of the loss, all four objectives use the same frozen Xiaomi readout features, 128-unit diagonal covariance readout, and 6,000/600/600 fitting, tuning and evaluation contexts specified in Appendix~\ref{app:context-sampling}.
Targets use the same 384 teacher samples per context, and all 600 evaluation contexts are retained.
Each objective is fitted with seeds 17 and 29, pairing initialization and minibatch sampling within each seed.
Table~\ref{tab:robust-variance} selects each checkpoint by its own tuning objective.

\begin{table}[htbp]
    \small
    \setlength{\tabcolsep}{6pt}
    \caption{Xiaomi diagonal variance-loss ablation on 600 evaluation contexts, with fitting seeds reported separately.
        SD error is $100\|\sigma_\theta-\widehat\sigma\|_2/\|\widehat\mu\|_2$, shown as median [25th, 75th percentile]; lower is better.
        Tail ratio is median predicted/target variance among the largest 1\% of targets; one is exact.}
    \label{tab:robust-variance}
    \begin{center}
        \begin{tabular}{lrcc}
            \toprule
            Objective      & Seed & SD error (\%)     & Tail ratio \\
            \midrule
            Original       & 17   & 5.58 [3.61, 9.03] & 0.577      \\
                           & 29   & 5.29 [3.30, 8.67] & 0.581      \\
            Variance Huber & 17   & 3.16 [1.78, 5.64] & 0.410      \\
                           & 29   & 3.24 [1.76, 5.71] & 0.411      \\
            Log MSE        & 17   & 2.81 [1.49, 5.55] & 0.244      \\
                           & 29   & 2.85 [1.48, 5.42] & 0.248      \\
            Log Huber      & 17   & 2.71 [1.49, 5.39] & 0.250      \\
                           & 29   & 2.77 [1.47, 5.51] & 0.238      \\
            \bottomrule
        \end{tabular}
    \end{center}
\end{table}

\paragraph{Accuracy and upper-tail trade-off.}
With log Huber, the mean tail ratio across the two seeds falls from 0.58 to 0.24, and raw variance MSE increases by 29\%.
The loss therefore changes the balance between typical SD error and errors on large variances.

These fits have not been evaluated in control rollouts, so the ablation establishes improved typical moment prediction rather than improved task performance.
Figure~\ref{fig:gar-variance-calibration} and the policy comparisons use the original deployed readouts.

\paragraph{Fitting details.}
All fits use AdamW with learning rate $10^{-3}$, weight decay $10^{-2}$, batch size 64 and gradient clipping at norm one.
Tuning is evaluated every 1,000 updates; after 12,000 updates, four evaluations without improvement reduce the learning rate by a factor of 0.3, down to $10^{-6}$, and eight further evaluations without improvement at that rate stop fitting.
All eight fits reach this stopping condition after 40,000--50,000 updates.

\FloatBarrier
\section{Task performance}
\label{app:performance}

\subsection{Inference cost of the moment readouts}
\label{app:gar-inference-cost}

To measure how much a single action-transformer evaluation saves in practice, we time the full policy call, including vision-language processing.
GAR is $1.91\times$ faster for Xiaomi and $1.57\times$ for GR00T on the main system (Table~\ref{tab:gar-inference-cost}).

\begin{table}[htbp]
    \small
    \caption{Batch-one latency of diagonal policies on two RTX 4090 systems, including vision-language processing.
        Mean and command readouts use 6,000 contexts; covariance fitting counts are shown separately.
        Speedup is teacher/GAR latency; lower latency and higher speedup are better.
        Three-session measurements use 64 calls per policy per session on 32 matched contexts and report medians of session means and speedups, with observed speedup ranges in brackets.
        The separate EPYC protocol uses 400 calls per policy on fifty contexts after ten warmups.}
    \label{tab:gar-inference-cost}
    \begin{center}
        \begin{tabular}{llrrr}
            \toprule
            Model & \shortstack{Covariance\\contexts} & Teacher (ms) & GAR (ms) & Speedup \\
            \midrule
            \multicolumn{5}{l}{\textit{Threadripper PRO 5955WX: three sessions}} \\
            Xiaomi & 6,000  & 220.25 & 115.36 & $\mathbf{1.91}\times$ [1.909, 1.915] \\
            GR00T  & 6,000  & 68.01  & 43.30  & $\mathbf{1.57}\times$ [1.566, 1.571] \\
            \midrule
            \multicolumn{5}{l}{\textit{EPYC 9554: earlier three sessions}} \\
            Xiaomi & 24,000 & 202.08 & 111.84 & $1.80\times$ [1.797, 1.807] \\
            GR00T  & 6,000  & 60.87  & 42.98  & $1.42\times$ [1.407, 1.421] \\
            \midrule
            \multicolumn{5}{l}{\textit{EPYC 9554: separate 400-call protocol}} \\
            Xiaomi & 6,000 & 221.31 & 132.92 & $1.66\times$ \\
            GR00T  & 6,000 & 59.77  & 41.74  & $1.43\times$ \\
            \bottomrule
        \end{tabular}
    \end{center}
\end{table}

Full and diagonal readouts have similar measured latency; variation across hosts and sessions is larger than their difference.

For Xiaomi, GAR reduces action-transformer time by replacing five evaluations with one (Table~\ref{tab:gar-inference-components}).
Vision-language processing takes similar time for both policies and accounts for approximately 29\% of teacher latency and 55\% of GAR latency in this profile.

\begin{table}[htbp]
    \small
    \caption{Xiaomi latency breakdown on the Threadripper PRO 5955WX / RTX 4090 system, using the 6,000-context mean and diagonal covariance readouts.
        Values are means over ten matched observations in a separate profiling session.
        Module times use CUDA events; total call latency uses synchronized wall-clock timing and also includes preprocessing, transfers and decoding.}
    \label{tab:gar-inference-components}
    \begin{center}
        \begin{tabular}{lrr}
            \toprule
            Component                   & Teacher (ms)   & GAR (ms)       \\
            \midrule
            Vision-language processing  & 65.4           & 66.8           \\
            Action transformer          & 135.4          & 28.3           \\
            GAR readout                 & ---            & 0.5            \\
            \midrule
            \textbf{Total call latency} & \textbf{225.9} & \textbf{120.7} \\
            \bottomrule
        \end{tabular}
    \end{center}
\end{table}

\paragraph{Timing protocol.}
The main timing comparison uses the exact 6,000-context mean, diagonal covariance and command readouts from Table~\ref{tab:gar_success} for both teachers.
Three independent sessions each use 32 matched held-out contexts, five warmup calls per policy and two timed repetitions per context, with policy order rotated.
Teacher and student share an RTX 4090 and an AMD Threadripper PRO 5955WX host (NVIDIA driver 550.120); CUDA is synchronized before and after each call.
Timing starts from in-memory observations and includes preprocessing, transfers, vision-language processing, action generation and decoded CPU actions.
Model loading, communication, queueing and simulation are excluded.

\subsection{Uncertainty in task-success differences}
\label{app:uncertainty}

\paragraph{Differences from the teacher.}
For both teachers, the intervals for the 6,000-context diagonal readout include zero but also allow losses of several percentage points; they do not establish practical equivalence.

Table~\ref{tab:gar-success-differences} reports success differences for every readout in Table~\ref{tab:gar_success}.
We independently resample the fifty binary episode outcomes within each task and policy, retaining equal weights across the fifty tasks, and repeat this 20,000 times.
Brackets give the 2.5th and 97.5th percentiles of the policy-minus-teacher differences.
Native scenes are unpaired, so episodes are not paired across policies by their indices.
These pointwise intervals describe repeated-episode uncertainty for the fixed task suite and fitted checkpoints; they omit fitting-seed and new-task variation and are not adjusted for multiple comparisons.
\begin{table}[htbp]
    \small
    \setlength{\tabcolsep}{4pt}
    \caption{Success differences on fifty tasks with fifty episodes per policy and task (2,500 total), using the settings of Table~\ref{tab:gar_success}.
        Teachers succeed on 1,456 Xiaomi and 577 GR00T episodes.
        Difference is readout minus teacher success in percentage points; higher favors the readout.
        Pointwise 95\% intervals use 20,000 independent within-task episode-bootstrap resamples (seed 85017), conditional on these tasks and fitted checkpoints.}
    \label{tab:gar-success-differences}
    \begin{center}
        \begin{tabular}{lllr rr}
            \toprule
            Teacher & Fitting contexts & Readout   & Successes & Difference & 95\% interval   \\
            \midrule
            Xiaomi  & 6,000            & Mean only & 1,416     & $-1.60$    & $[-3.84,0.60]$  \\
                    &                  & Diagonal  & 1,441     & $-0.60$    & $[-2.84,1.60]$  \\
                    &                  & Full      & 1,432     & $-0.96$    & $[-3.20,1.20]$  \\
            \addlinespace
                    & 24,000           & Mean only & 1,411     & $-1.80$    & $[-4.08,0.40]$  \\
                    &                  & Diagonal  & 1,443     & $-0.52$    & $[-2.72,1.68]$  \\
                    &                  & Full      & 1,411     & $-1.80$    & $[-4.04,0.40]$  \\
            \midrule
            GR00T   & 6,000            & Mean only & 493       & $-3.36$    & $[-5.24,-1.52]$ \\
                    &                  & Diagonal  & 548       & $-1.16$    & $[-3.04,0.68]$  \\
                    &                  & Full      & 508       & $-2.76$    & $[-4.64,-0.88]$ \\
            \addlinespace
                    & 24,000           & Mean only & 505       & $-2.88$    & $[-4.76,-1.00]$ \\
                    &                  & Diagonal  & 536       & $-1.64$    & $[-3.52,0.24]$  \\
                    &                  & Full      & 529       & $-1.92$    & $[-3.80,-0.04]$ \\
            \bottomrule
        \end{tabular}
    \end{center}
\end{table}

\paragraph{Diagonal sampling versus mean-only execution.}
To isolate the contribution of continuous-action sampling, each comparison keeps the learned mean and stochastic command predictor fixed.
All four observed success differences favor diagonal sampling, but only GR00T with 6,000 fitting contexts has a pointwise 95\% interval entirely above zero (Table~\ref{tab:gar-diagonal-mean-differences}).
We use the same independent within-task bootstrap as above.

\begin{table}[htbp]
    \small
    \centering
    \caption{Diagonal-minus-mean success differences in percentage points, with pointwise 95\% intervals from 20,000 independent within-task bootstrap resamples (seed 85017).
        Positive differences favor diagonal sampling; each policy has fifty episodes per task on the fixed fifty-task suite.}
    \label{tab:gar-diagonal-mean-differences}
    \begin{tabular}{llrr}
        \toprule
        Teacher & Fitting contexts & Difference & 95\% interval  \\
        \midrule
        Xiaomi  & 6,000            & $+1.00$    & $[-1.20,3.24]$ \\
                & 24,000           & $+1.28$    & $[-0.96,3.52]$ \\
        \midrule
        GR00T   & 6,000            & $+2.20$    & $[0.40,3.96]$  \\
                & 24,000           & $+1.24$    & $[-0.56,3.04]$ \\
        \bottomrule
    \end{tabular}
\end{table}

\subsection{Conditional versus fixed covariance}
\label{app:fixed-covariance}

GAR's predicted covariance gives closer translation and rotation action-chunk distributions to the teacher than a fixed covariance, while its effect on task success remains uncertain.
We compare the 6,000-context diagonal GAR with a \emph{Fixed} policy that replaces its context-dependent covariance by the average predicted covariance over the 6,000 fitting contexts.
Both policies retain identical learned means and stochastic command predictors, and use variance multiplier one.
The fixed covariance is computed from the fitted GAR readout, so this baseline requires the same context collection, teacher sampling and readout fitting as GAR.

\paragraph{Task success.}
GAR has higher observed success for both teachers, but the intervals for GAR-minus-Fixed include zero (Table~\ref{tab:gar-fixed-success}).
Each policy uses fifty episodes on each of the fifty RoboCasa365 tasks, with independently drawn scenes and the execution settings of Table~\ref{tab:gar_success}.

The Fixed evaluation combines five earlier episodes per task with forty-five additional episodes, without selecting a variance scale.
To check the effect of reusing the earlier episodes, we also restrict both policies to the latter episode indices.
This gives differences of $+1.91$ percentage points for Xiaomi and $+1.11$ for GR00T, with 95\% intervals $[-0.36,4.18]$ and $[-0.80,3.02]$, respectively.

\begin{table}[htbp]
    \small
    \centering
    \setlength{\tabcolsep}{6pt}
    \caption{GAR versus Fixed success on RoboCasa365, with 2,500 episodes per policy and teacher.
        Both policies share their mean and command readouts; only the covariance changes.
        Success is a percentage, and differences are GAR minus Fixed in percentage points; higher favors GAR.
        Brackets give pointwise 95\% independent within-task episode-bootstrap intervals, conditional on the fitted checkpoints and fixed task suite.}
    \label{tab:gar-fixed-success}
    \begin{tabular}{lrrrr}
        \toprule
        Teacher & GAR   & Fixed & Difference & 95\% interval  \\
        \midrule
        Xiaomi  & 57.64 & 55.92 & $+1.72$    & $[-0.44,3.92]$ \\
        GR00T   & 21.92 & 20.76 & $+1.16$    & $[-0.64,3.00]$ \\
        \bottomrule
    \end{tabular}
\end{table}

\paragraph{Distribution fit.}
Task success summarizes the outcome of an execution; we also compare the action distributions that produce it.
GAR has lower mean distribution error in all four comparisons, with individual 95\% paired context-bootstrap intervals excluding zero (Table~\ref{tab:gar-fixed-fit}).
This supports more faithful translation and rotation chunk distributions from conditional covariance, although the task-success comparison does not establish a corresponding control benefit.
The distances do not assess translation--rotation coupling, other continuous channels or binary commands.

We compare the deployed readouts directly with teacher samples on 143 held-out contexts per teacher, excluding contexts used for fitting or checkpoint selection.
At each context, the final 512 samples from the FP32 teacher banks in Appendix~\ref{app:physical_action_scale} provide the reference distribution.
We evaluate the saved readouts in FP32 on cached readout features and measure sliced 1-Wasserstein distance separately for the 48 translation and 48 rotation coordinates of each sixteen-action chunk.
Both policies use the same clipping and physical scaling, 256 projection directions and 2,048 standard-normal Sobol reference points, following Appendix~\ref{app:chunk-physical-action-scale}.
No Gaussian is fitted to the evaluation samples.

\begin{table}[htbp]
    \small
    \centering
    \setlength{\tabcolsep}{4pt}
    \caption{Mean sliced 1-Wasserstein distance to teacher action chunks on 143 held-out contexts per teacher; lower is better.
        GAR and Fixed share their means and command predictors.
        Distances and reductions use the action group's stated unit; reduction is Fixed minus GAR, so positive values favor GAR.
        Brackets give pointwise 95\% paired context-bootstrap intervals from 10,000 resamples, conditional on the fitted readouts and cached samples.}
    \label{tab:gar-fixed-fit}
    \begin{tabular}{llrrr}
        \toprule
        Teacher & Action group (unit) & GAR   & Fixed & Reduction [95\% interval] \\
        \midrule
        Xiaomi  & Translation (mm)    & 0.566 & 0.987 & $0.420\;[0.359,0.483]$    \\
                & Rotation (degrees)  & 0.145 & 0.196 & $0.051\;[0.042,0.061]$    \\
        \addlinespace
        GR00T   & Translation (mm)    & 1.560 & 2.647 & $1.088\;[0.887,1.299]$    \\
                & Rotation (degrees)  & 0.336 & 0.636 & $0.301\;[0.259,0.343]$    \\
        \bottomrule
    \end{tabular}
\end{table}

\FloatBarrier
\subsection{Gaussian approximation and sampling}
\label{app:gr00t-recovery}

\paragraph{Policies and native evaluation.}
The controls in Section~\ref{sec:experiments_diagnostics} evaluate the original 6,000-context diagonal readouts on all sixteen seen composite tasks.
They use the same native evaluator, observation histories, task horizons and sixteen-action execution as Table~\ref{tab:gar_success}.
Episodes begin with fresh native resets, without pairing physical starts across policies.
The five intervention policies each use 32 episodes per task, giving 512 episodes per policy and model.

The GAR, mean-only GAR and raw one-step references reuse the 800 episodes per policy from the same sixteen tasks in Table~\ref{tab:gar_success}, with 50 episodes per task.
Their checkpoints and execution rules are unchanged.
Raw one-step uses native command decoding; mean-only GAR retains GAR commands.
The native teacher in that table uses its own joint motion and command distribution; the controls here retain GAR commands.
For GR00T, teacher sampling with GAR commands scores 11.91\% in Table~\ref{tab:gar_diagnostics}, whereas the native teacher scores 15.38\% in Table~\ref{tab:gar_success}.
These policies also use independent episode draws, so their difference does not isolate the effect of command handling.

At every observation, a bank of 256 samples from the deployed teacher, using four steps for GR00T and five for Xiaomi, defines three replacement policies.
The diagonal Gaussian uses the bank mean and marginal variances; the full Gaussian also retains all cross-coordinate and temporal covariances; empirical motion selects one complete continuous-action chunk uniformly from the bank.
The covariance uses denominator 256, so the full Gaussian has exactly the mean and covariance of this finite empirical distribution.
These quantities are matched before the common controller transformations.
They estimate the deployed coarse sampler's moments, rather than the continuous-field moments in Section~\ref{sec:gar_flow_moments}.
All three replacements retain the original GAR command predictor and sampling rule at the current observation.
Repeated teacher sampling is an expensive diagnostic, not part of deployed GAR inference.

For Xiaomi, we check the batched teacher output against singleton evaluation at a fixed noise input at every observation.
Starting from batch size 32, we reduce it through 16, 8, 4, 2 and 1 until outputs agree within relative tolerance 0.02 and absolute tolerance 0.015625; singleton evaluation must agree exactly.
All reported Xiaomi episodes use this check at every observation.

\begin{table}[htbp]
    \small
    \centering
    \caption{Native-protocol diagnostic outcomes across sixteen seen composite tasks, in successes/episodes.
        All policies use GAR's binary-command rule.
        The reference reuses 50 episodes per task from Table~\ref{tab:gar_success}; each intervention uses 32 per task.}
    \label{tab:gar-motion-replacements}
    \begin{tabular}{lrr}
        \toprule
        Continuous-motion policy & GR00T & Xiaomi \\
        \midrule
        GAR: predicted diagonal moments & 94 / 800 & 464 / 800 \\
        Gaussian: teacher-estimated diagonal moments & 67 / 512 & 300 / 512 \\
        Gaussian: teacher-estimated full moments & 60 / 512 & 304 / 512 \\
        Teacher sampling, GAR commands & 61 / 512 & 296 / 512 \\
        Teacher-estimated mean, no motion noise & 58 / 512 & 283 / 512 \\
        Teacher-estimated mean, GAR variances & 75 / 512 & 308 / 512 \\
        \bottomrule
    \end{tabular}
\end{table}

\paragraph{Uncertainty and interpretation.}
Success rates give each task equal weight.
For each contrast, we independently resample episodes within each task and policy 4,000 times and report the 2.5th and 97.5th percentiles of the difference in task-averaged success.
These pointwise intervals describe episode variability for the fixed tasks and readouts, not variability across new tasks or readout fits.

The empirical-motion comparison tests structure beyond the teacher bank's mean and full covariance under a common command rule.
Neither model shows a clear empirical-motion advantage, but these intervals do not establish exact Gaussianity or practical equivalence.
Replacing both the predicted mean and variances gives small, uncertain gains for both models.

\paragraph{Correcting the mean.}
We isolate mean prediction by replacing only the mean, with either no motion noise or GAR's predicted variance, and keep GAR's command rule unchanged.
For GR00T without motion noise, success rises from 62/800 (7.75\%) to 58/512 (11.33\%), a gain of 3.58 points $[0.53,6.84]$.
With GAR's predicted variances, it rises from 94/800 (11.75\%) to 75/512 (14.65\%), a gain of 2.90 points $[-0.52,6.49]$.
This supports mean prediction as a limitation of GR00T's mean-only policy; the gain under Gaussian sampling remains uncertain.
For Xiaomi without motion noise, mean substitution changes success from 58.13\% to 55.27\%, or $-2.85$ points $[-7.61,1.79]$.
With GAR variances, success changes from 58.00\% to 60.16\%, or $+2.16$ points $[-2.48,6.95]$.

\paragraph{Sampling around the teacher-estimated mean.}
\label{app:gar-sampling-controls}
Both sampling-control policies use the mean of 256 teacher samples at each observation and retain GAR's stochastic binary-command rule.
One executes this mean without continuous-motion noise; the other samples Gaussian motion using GAR's predicted diagonal covariance.
Thus, the comparison changes motion sampling while holding the mean estimator and command rule fixed.
Observed success increases for both models, but both intervals narrowly include zero (Table~\ref{tab:gar_diagnostics}).

\subsection{LIBERO-Long}
\label{app:libero-long}

LIBERO-Long tests whether GAR retains near-teacher performance beyond RoboCasa365.
Across its ten tasks, GAR succeeds in 477/500 episodes, compared with 483/500 for the teacher and 480/500 for raw one-step execution (Table~\ref{tab:libero-long-success}).
Appendices~\ref{app:libero-setting} and~\ref{app:libero-fitting} specify evaluation and fitting.
Table~\ref{tab:libero-long-success} reports successes, with fifty episodes in each task row.

\begin{table}[htbp]
    \small
    \centering
    \setlength{\tabcolsep}{5pt}
    \caption{LIBERO-Long successes on fifty common initial states per task.
        The teacher is \mbox{Xiaomi-Robotics-0}; diagonal GAR uses 6,000 fitting contexts.
        Overall entries are successes out of 500 episodes.}
    \label{tab:libero-long-success}
    \begin{tabular}{lrrr}
        \toprule
        Task                                 & Teacher          & Raw one-step     & GAR              \\
        \midrule
        Soup and tomato sauce into basket    & 49               & 47               & 48               \\
        Cream cheese and butter into basket  & 48               & 47               & 47               \\
        Turn on stove and place moka pot     & 50               & 50               & 50               \\
        Bowl into drawer and close it        & 48               & 46               & 48               \\
        Two mugs onto plates                 & 50               & 50               & 50               \\
        Book into caddy                      & 50               & 50               & 50               \\
        Mug onto plate and pudding beside it & 44               & 45               & 41               \\
        Soup and cream cheese into basket    & 50               & 50               & 50               \\
        Two moka pots onto stove             & 45               & 47               & 45               \\
        Mug into microwave and close it      & 49               & 48               & 48               \\
        \midrule
        \textbf{Overall}                     & \textbf{483/500} & \textbf{480/500} & \textbf{477/500} \\
        \bottomrule
    \end{tabular}
\end{table}

\FloatBarrier
GAR's success difference is $-1.2$ percentage points relative to the teacher, with a paired 95\% interval of $[-3.0,+0.6]$, and $-0.6$ points relative to raw one-step, with an interval of $[-2.2,+1.0]$.
We compute these intervals using 10,000 bootstrap resamples of paired episodes within each task, retaining equal task weights.
The intervals condition on these ten tasks and one fitted checkpoint.

\section{Behavioral diversity}
\label{app:diversity}

\subsection{Constructed Gaussian policy}
\label{app:toy-gaussian}

The hand-designed policy in Figure~\ref{fig:gaussian-behavior} illustrates how Gaussian actions can produce two routes through state feedback.
Near the start, the lateral mean amplifies small positive or negative deviations from $y=0$, separating trajectories around the obstacle.
Past the obstacle, the mean draws paths back toward $y=0$ and the goal $(3,0)$.
The same policy thus follows either route while using a single Gaussian at every state.

For state $s_k=(x_k,y_k)$, actions update position directly:
\[
    a_k\sim\mathcal N\!\left(\mu(s_k),\Sigma\right),
    \qquad s_{k+1}=s_k+a_k,
    \qquad \Sigma=\operatorname{diag}(0.025^2,0.035^2).
\]
The mean depends only on the current state:
\[
    \begin{aligned}
        \mu_x(x,y) & =0.08\tanh\!\left(1.8(3-x)\right),                 \\
        \mu_y(x,y) & =0.20\left[0.92g(x)\tanh(5y)-y\right],             \\
        g(x)       & =\left[1+\exp\!\left(5(x-0.95)\right)\right]^{-1}.
    \end{aligned}
\]
As $g(x)$ decreases, the lateral mean changes from separating the paths to bringing them together.

This position update connects Gaussian action sampling to diffusion in physical time: it is an Euler--Maruyama step, with unit step size, for
\begin{equation}
    ds_\tau=\mu(s_\tau)\,d\tau+\Sigma^{1/2}\,dW_\tau,
    \label{eq:toy-physical-diffusion}
\end{equation}
where $\tau$ denotes physical time and $W_\tau$ is standard two-dimensional Brownian motion.
This interpretation applies to the constructed position model.

All 128 rollouts start at $(-3,0)$ and execute 110 Gaussian actions, using NumPy's default random generator with seed 2026.
The circle of radius 0.5 at the origin marks the obstacle; the Gaussian update is applied unchanged throughout.
Every generated trajectory is displayed, with colors assigned afterward by the sign of $y_{38}$.

\subsection{Diagonal behavioral diversity}
\label{app:task-entropy}

We next test whether GAR produces different object-movement orders from the same initial scene, and compare its behavioral diversity with the teacher's.
We use the Xiaomi 6,000-context Diagonal policy from the main benchmark, with its predicted variance and independent Bernoulli commands.
The comparison covers two scenes each for loading a dishwasher, packing identical lunches and storing leftovers in a bowl, with 32 executions per policy and scene (384 episodes in total).
These scenes were selected during exploratory search and do not estimate the prevalence of alternatives across the benchmark.

To hold the starting context fixed, both policies receive the same saved initial images, robot-state history and instruction.
Both use the same evaluation runtime, native preprocessing and decoding, task horizons and sixteen-action replanning.
For each policy and scene, a fixed-seed check confirms that actions and their sixteen-action physical replay are repeatable; these twelve checks are separate from the 384 episodes.

\paragraph{Different orders in successful executions.}
\label{app:successful-orders}
In the second lunch-packing scene, Diagonal succeeds in 3/32 attempts with two movement orders (Figure~\ref{fig:robocasa-successful-orders}).
Two successes move garlic A, garlic B, chicken B and chicken A; the third moves garlic A, chicken A, garlic B and chicken B.
All three order labels remain unchanged at 3, 5 and 8 cm displacement thresholds.
The teacher also succeeds in 3/32 attempts, with three distinct orders; only one matches an observed successful Diagonal order.

These executions demonstrate distinct successful behaviors under a Gaussian continuous-action policy.
The policy also samples Bernoulli commands, so the experiment does not isolate continuous-action noise as the cause of the different orders.

Across the six starts, Diagonal succeeds in 13/192 attempts and the teacher in 12/192.
Both policies succeed nine times in the second leftovers scene, always with the same observed order; Diagonal also succeeds once in the first lunch-packing scene, where its order label varies with the displacement threshold.
Neither policy succeeds in the other three scenes.

\paragraph{Choice and order entropy.}
The successful examples establish that different orders are possible; we now compare how much behavior varies across all attempts, including failures.
First-choice entropy measures variation in which object moves first, while full-order entropy measures variation in the sequence of objects moved.
For category frequencies $\hat p_k$, Shannon entropy $-\sum_k\hat p_k\log_2\hat p_k$ measures diversity in bits, and is zero when only one category is observed.

\begin{figure}[htbp]
    \centering
    \includegraphics[width=\linewidth]{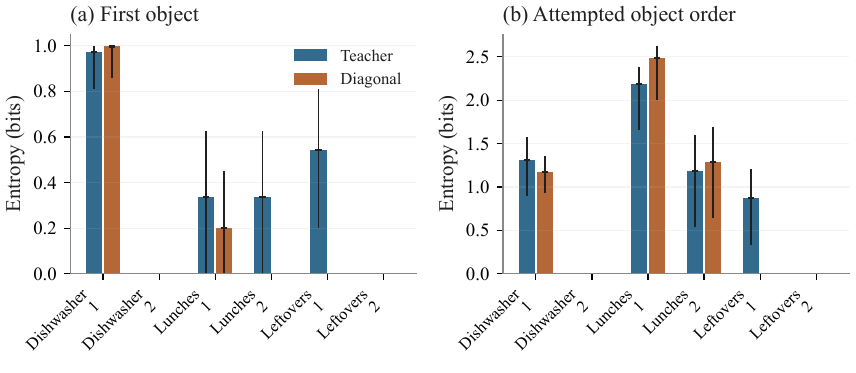}
    \caption{Behavioral diversity of diagonal GAR and the Xiaomi teacher from identical starts.
        Each scene has 32 executions per policy, including all successes and failures.
        Bars show first-object and full movement-order entropy at 5 cm; lines show 95\% within-policy, within-scene bootstrap intervals.
        Zero observed entropy does not exclude rare alternatives, and equal entropy does not establish matching choices.}
    \label{fig:diagonal-task-entropy}
\end{figure}

An object counts as moving after exceeding 5 cm displacement for three consecutive physical steps; thresholds of 3 and 8 cm test sensitivity.
First-choice categories include ties and no movement, and full-order categories include incomplete attempts.
We compute entropy over all 32 attempts per policy and scene, and give the six scene entropies equal weight.
Figure~\ref{fig:diagonal-task-entropy} shows the scene entropies, and Table~\ref{tab:diagonal-first-choice-counts} gives the first-object counts.

\begin{table}[htbp]
    \caption{First-object counts for diagonal GAR and the Xiaomi teacher across all 32 attempts per policy and scene, using the 5 cm threshold.
        These are movement choices rather than successful completions; no ties or no-movement outcomes occur in these scenes.
        A and B distinguish objects of the same type within a scene.}
    \label{tab:diagonal-first-choice-counts}
    \small
    \begin{center}
        \begin{tabularx}{\linewidth}{@{}l>{\raggedright\arraybackslash}X>{\raggedright\arraybackslash}X@{}}
            \toprule
            Scene        & Teacher counts                      & Diagonal counts               \\
            \midrule
            Dishwasher 1 & cup: 19; bowl: 13                   & cup: 15; bowl: 17             \\
            Dishwasher 2 & dish A: 32                          & dish A: 32                    \\
            Lunches 1    & vegetable A: 2; vegetable B: 30     & vegetable A: 1; vegetable B: 31 \\
            Lunches 2    & chicken A: 2; garlic A: 30          & garlic A: 32                  \\
            Leftovers 1  & chicken drumstick: 4; vegetable: 28 & vegetable: 32                 \\
            Leftovers 2  & vegetable: 32                       & vegetable: 32                 \\
            \bottomrule
        \end{tabularx}
    \end{center}
\end{table}

Diagonal produces less varied first-object choices on these scenes.
Mean first-choice entropy is 0.365 bits for the teacher and 0.200 bits for Diagonal, a Diagonal-minus-teacher difference of $-0.166$ bits with a 95\% within-scene bootstrap interval of $[-0.266,-0.043]$.
Diagonal has lower first-choice entropy in three scenes, slightly higher entropy in the first dishwasher scene, and zero observed entropy for both policies in the other two scenes.

For full movement orders, mean entropy is 0.925 bits for the teacher and 0.824 bits for Diagonal, with a difference of $-0.101$ bits and interval $[-0.282,0.088]$.
This interval includes zero and does not establish equal entropy.

The first-choice reduction persists when the movement threshold changes: at 3 and 8 cm, differences remain negative at $-0.225$ and $-0.110$ bits, with intervals $[-0.313,-0.101]$ and $[-0.195,-0.006]$.
All three full-order difference intervals include zero.
Resampling is independent within each policy and scene; intervals condition on the observed categories and exclude unseen alternatives, scene-selection uncertainty and variation across new scenes.

\clearpage
\section{Limitations}
\label{app:limitations}

\begin{itemize}
    \item \textbf{Evaluation scope.}
          Our experiments evaluate selected pretrained policies on RoboCasa365 and LIBERO-Long, both simulation benchmarks.
          Generalization to physical robots, other embodiments and other pretrained policies remains untested.
          The task-success intervals describe episode variability for the evaluated tasks and fitted checkpoints; they exclude variation across readout fits and new tasks, and do not establish practical equivalence to the teachers (Appendices~\ref{app:uncertainty} and~\ref{app:libero-long}).

    \item \textbf{Distributional approximation.}
          Near-teacher task success does not imply that GAR reproduces the teacher's full action distribution.
          Diagonal GAR omits correlations between continuous coordinates and action positions, while both covariance variants assume binary commands are mutually independent and independent of continuous actions given the context.
          Our Gaussian-approximation measurements assess translation and rotation separately, leaving their joint dependence and other command groups outside the analysis (Appendix~\ref{app:physical_action_scale}).

    \item \textbf{Moment-prediction errors.}
          Moment prediction remains imperfect, with substantial errors at some contexts and underestimation of the largest variances, especially for Xiaomi (Appendix~\ref{app:moment-diagnostics}).
          Readouts fitted on demonstration contexts may also encounter different states during execution.
          Correcting the mean improves GR00T's mean-only success, but its benefit under GAR sampling remains uncertain (Appendix~\ref{app:gr00t-recovery}).
          Whether errors at particular execution states explain the remaining performance gap remains unresolved.

    \item \textbf{Behavioral diversity.}
          Gaussian action distributions can produce multimodal trajectories through state feedback, as our constructed example demonstrates, and GAR produces distinct successful object-movement orders.
          Preservation of the teacher's behavioral diversity is not established: across six selected scenes, first-object choice entropy is lower for GAR, while the difference in full movement-order entropy is inconclusive (Appendix~\ref{app:task-entropy}).
          These measurements include unsuccessful attempts and do not establish how frequently diverse successful behaviors occur across the benchmark.

    \item \textbf{Preparation and inference cost.}
          Fitting small readouts is inexpensive once teacher statistics are cached, but preparing those statistics requires repeated teacher sampling at each context (Appendix~\ref{app:fitting-cost}).
          The small fraction of fitting contexts therefore does not describe the full distillation cost.
          At inference, GAR retains the teacher's vision-language processing, which limits the total speedup from replacing iterative action generation (Appendix~\ref{app:gar-inference-cost}).
\end{itemize}

\end{document}